%% file: main.tex
\pdfoutput=1 
\documentclass{article}
\usepackage{iclr2027_conference,times}

\input{math_commands.tex}

\usepackage[utf8]{inputenc} 
\usepackage[T1]{fontenc}    
\usepackage{hyperref}       
\usepackage{url}            
\usepackage{booktabs}       
\usepackage{amsfonts}       
\usepackage{amsmath}        
\usepackage{amssymb}        
\usepackage{nicefrac}       
\usepackage{microtype}      
\usepackage{xcolor}         
\usepackage{graphicx}       
\usepackage{eqparbox}
\usepackage{algorithm}
\usepackage[noend]{algpseudocode}
\usepackage{comment}
\usepackage{subcaption}
\usepackage{wrapfig}
\usepackage{multirow}
\usepackage{array}
\algrenewcommand{\algorithmiccomment}[1]{\hfill\textcolor{gray}{$\triangleleft$ \textit{#1}}}

\title{\raggedright\hyphenpenalty=10000\exhyphenpenalty=10000
DPMPC: Learning Diffusion Priors for Model Predictive Control
}

\author{%
  Minghui Ye \\
  Heriot-Watt University \\
  \texttt{my2035@hw.ac.uk} \\
  \And
  Ahmed H.~Elsheikh \\
  Heriot-Watt University \\
  \texttt{a.elsheikh@hw.ac.uk} \\
}

\iclrfinalcopy
\begin{document}

\maketitle
\lhead{Preprint}

\begin{abstract}

Model predictive control (MPC) has achieved notable success in continuous control.
MPC often uses a policy prior to propose candidate trajectories and to learn the value function.
When the prior differs from the executed planner, this \emph{actor divergence} biases the value estimates of candidate trajectories.
The commonly used Gaussian prior cannot represent the non-Gaussian action distributions induced by stochastic planning, leaving a residual divergence.
To address this, we propose DPMPC (Diffusion Priors for Model Predictive Control), which trains an expressive diffusion prior to match the planner's action distribution.
Generalized lazy reanalyze estimates this distribution from multiple recent MPC realizations stored per replay state, without rerunning MPC at every policy update.
An advantage-weighted score-matching objective then fits the prior to this estimate while emphasizing high-value actions.
DPMPC achieves state-of-the-art asymptotic performance and sample efficiency across 14 continuous-control tasks from DMControl, MyoSuite, and HumanoidBench.
Further analysis shows that DPMPC yields more accurate value estimates than Gaussian-prior baselines.
Code is available at \url{https://github.com/yeminghui/dpmpc}.

\end{abstract}

\section{Introduction}

Reinforcement learning (RL) has achieved impressive results in domains spanning robotics~\citep{zitkovich2023rt,liang2025residual}, games~\citep{schrittwieser2020mastering, hafner2025mastering}, and scientific and engineering applications~\citep{mankowitz2023faster, lagemann2026hydrogym}. 
Yet these successes often require large amounts of interaction data, which hinders real-world deployment~\citep{ye2025model,wu2023daydreamer}. 
Planning-driven model-based reinforcement learning (MBRL), which combines a learned world model with planning, has emerged as one of the most sample-efficient paradigms~\citep{schrittwieser2020mastering, Hansen2022tdmpc}.

Planning-driven MBRL methods typically run model predictive control (MPC) over a short horizon and score each trajectory with a terminal value from a learned value function~\citep{Hansen2022tdmpc,hansen2023tdmpc2}.
Planning performance thus depends directly on the accuracy of this value.
The value function is learned with bootstrapped targets whose actions come from a policy prior, the same network that proposes candidate actions for MPC.
It therefore estimates the value of the prior rather than that of the executed planner.
The mismatch between the prior and the planner's action distribution, known as \emph{actor divergence}~\citep{sikchi2021learning,zhan2025boom}, directly biases the value estimate.
Recent methods reduce this divergence by aligning a Gaussian prior with the planner, either through imitation~\citep{wang2025bootstrapped} or likelihood maximization~\citep{zhan2025boom}.
Sampling-based MPC, however, solves a non-convex problem: independent runs from the same state can reach different high-return actions, so the planner's action distribution is a non-Gaussian, potentially multimodal mixture~\citep{sasaki2021variational,zhan2025boom}.
A Gaussian prior cannot represent this mixture, leaving a residual actor divergence that biases the value estimate.

This limitation motivates more expressive policy priors based on diffusion models~\citep{sohl2015deep,ho2020ddpm,song2021scorebased}, which have proven effective at capturing multimodal action distributions in imitation learning~\citep{chi2025diffusion,reuss2023goal} and offline RL~\citep{wang2023diffusion}.
These successes, however, depend on a fixed dataset whose actions directly serve as denoising targets.
In online RL, no such dataset exists: the target distribution is unknown~\citep{ma2025efficient}.
The few online methods therefore rely on a learned critic, either backpropagating value gradients through the denoising chain~\citep{wang2024diffusion} or optimizing critic-guided surrogate objectives~\citep{psenka2024learning,ding2024diffusion,ma2025efficient,dong2025maximum}.
Because the target is specified only implicitly through a nonstationary value estimate, training can be costly, unstable, or sample-inefficient.
Moreover, these methods are predominantly model-free, and diffusion policy priors for planning-driven MBRL remain largely unexplored.

Our key insight is that, in planning-driven MBRL, the planner itself can supply an explicit training target: its action distribution, which is exactly what the prior must match to reduce \emph{actor divergence}.
Matching this distribution turns online diffusion policy training into a supervised learning problem.
Building on this insight, we propose \textbf{D}iffusion \textbf{P}riors for \textbf{M}odel \textbf{P}redictive \textbf{C}ontrol (\textbf{DPMPC}). 
A naive implementation, however, would rerun MPC to sample the planner distribution at every policy update, which is computationally prohibitive.
We address this with generalized lazy reanalyze, which stores multiple MPC realizations per replay state and periodically refreshes them to estimate the planner distribution.
We then train the diffusion policy prior on samples from this estimate with an advantage-weighted score-matching loss that shifts mass toward higher-value actions while keeping the prior close to the planner distribution.

In this paper, we make three key contributions: (1) We identify the residual \emph{actor divergence} left by Gaussian policy priors as a source of value bias in planning-driven MBRL.
We show that the value estimates of recent Gaussian-prior methods retain a large error relative to the planner's return throughout training.
(2) We develop a simple yet effective recipe for training diffusion policy priors.
It combines generalized lazy reanalyze, which maintains an estimate of the planner distribution, with an advantage-weighted score-matching loss that trains the prior toward high-value actions while keeping it close to this distribution.
(3) We demonstrate that DPMPC achieves state-of-the-art performance across 14 tasks from DMControl~\citep{tassa2018dmcontrol}, MyoSuite~\citep{MyoSuite2022}, and HumanoidBench~\citep{sferrazza2024humanoidbench}.
Our analysis shows that DPMPC provides more accurate value estimates than Gaussian-prior baselines.

\begin{figure}[t]
\centering
\includegraphics[width=0.95\textwidth]{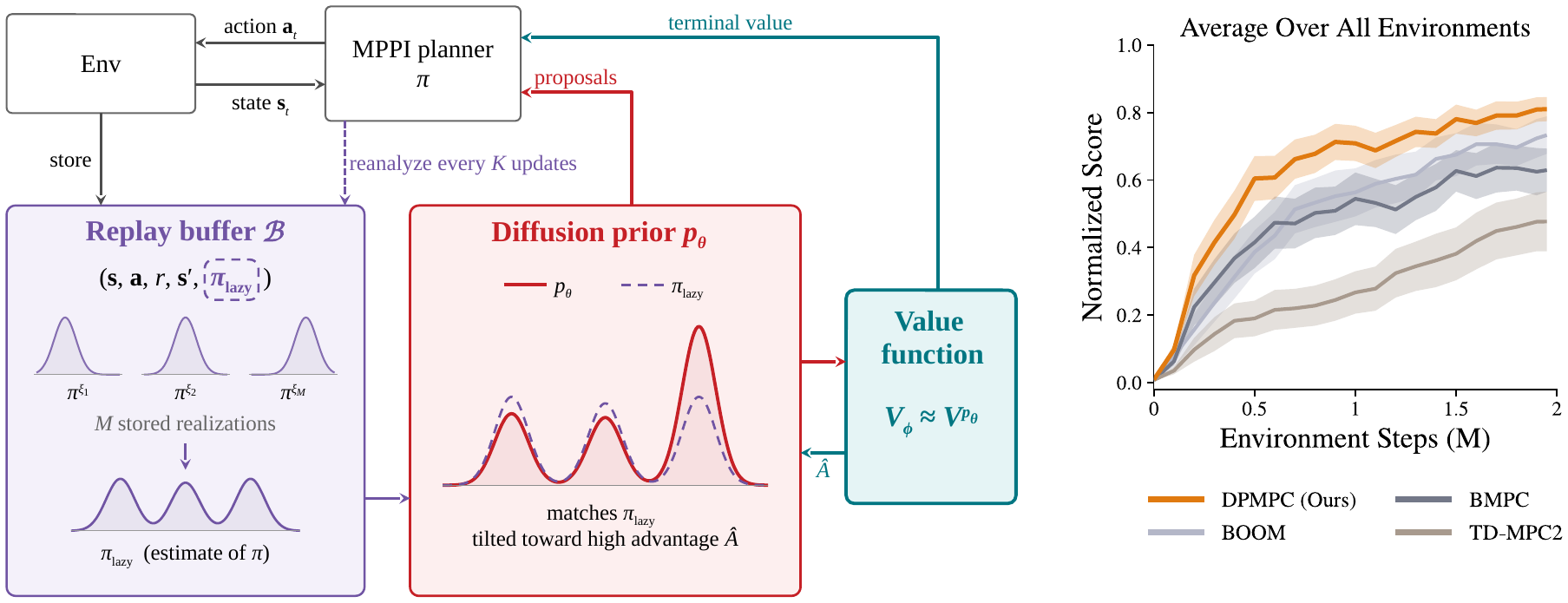}
\caption{\textbf{Left: Overview of DPMPC.}
The MPPI planner $\pi$ acts in the environment, guided by proposals from the diffusion prior $p_\theta$ and terminal values from $V_\phi$.
The prior is trained to match the planner distribution, estimated as a uniform mixture $\pi_{\text{lazy}}$ of $M$ MPC realizations stored in the replay buffer.
Because $V_\phi$ is learned under the prior ($V_\phi \approx V^{p_\theta}$), matching $p_\theta$ to $\pi$ makes it an accurate terminal value for the planner.
\textbf{Right: Average performance} over 14 tasks of DPMPC and baselines.}
\label{fig:dpmpc_loop}
\end{figure}

\section{Background}

We consider a Markov Decision Process (MDP)~\citep{puterman2014markov} defined by $(\mathcal{S}, \mathcal{A}, \mathcal{P}, \mathcal{R}, \gamma, \mu)$, where $\mathcal{S}$ is the state space, $\mathcal{A}$ is the action space, $\mathcal{P}: \mathcal{S} \times \mathcal{A} \rightarrow \Delta(\mathcal{S})$ is the transition dynamics,
$\mathcal{R}: \mathcal{S} \times \mathcal{A} \rightarrow \mathbb{R}$ is the reward function,
$\gamma \in (0,1]$ is the discount factor, and
$\mu \in \Delta(\mathcal{S})$ is the initial state distribution.
A policy $\pi: \mathcal{S} \rightarrow \Delta(\mathcal{A})$ defines a distribution over actions given states.
The objective is to find an optimal policy that maximizes the expected discounted return $\mathbb{E}_{s_0 \sim \mu,\, a_t \sim \pi(\cdot|s_t),\, s_{t+1} \sim \mathcal{P}(\cdot|s_t,a_t)} \left[ \sum_{t=0}^{\infty} \gamma^t \mathcal{R}(s_t, a_t) \right]$.

\subsection{Planning-driven MBRL}
\label{subsec:planning_based_mbrl}
Planning-driven MBRL methods~\citep{hansen2023tdmpc2,wang2025bootstrapped,zhan2025boom} combine a latent-space world model with MPC.
The agent does not execute the parametric policy directly; instead, it uses the policy as a prior $p(\mathbf{a}\mid \mathbf{z})$ to guide online planning.
An observation $\mathbf{s}$ is encoded into a latent state $\mathbf{z} = h(\mathbf{s})$, on which the learned dynamics $\mathbf{z}' = d(\mathbf{z},\mathbf{a})$ and reward $\hat r = R(\mathbf{z},\mathbf{a})$ are defined.
We refer to TD-MPC2~\citep{hansen2023tdmpc2} for world-model training details.

\paragraph{MPC with a policy prior.}
The planner is commonly instantiated with Model Predictive Path Integral (MPPI)~\citep{williams2015model}.
Given the current latent state $\mathbf{z}_t$, MPPI optimizes a Gaussian search distribution $\mathcal{N}(\boldsymbol{\mu}_{t:t+H}, \boldsymbol{\sigma}_{t:t+H}^2)$ over action sequences $\mathbf{a}_{t:t+H}$, where $H$ is the planning horizon.
The policy prior $p(\mathbf{a}\mid \mathbf{z})$ contributes additional proposal trajectories to the candidate set.
All candidates are rolled out through the latent dynamics and evaluated under the finite-horizon objective:
\begin{equation}
(\boldsymbol{\mu}_{t:t+H}^*, \boldsymbol{\sigma}_{t:t+H}^*)
= \arg\max_{\boldsymbol{\mu}, \boldsymbol{\sigma}} \;
\mathbb{E}_{\mathbf{a}_{t:t+H} \sim \mathcal{N}(\boldsymbol{\mu}, \boldsymbol{\sigma}^2)}
\!\left[
\sum_{h=t}^{t+H-1}\gamma^{\,h-t} R(\mathbf{z}_h,\mathbf{a}_h)
+ \gamma^H U(\mathbf{z}_{t+H}, \mathbf{a}_{t+H})
\right],
\label{eq:bmpc_planning_objective}
\end{equation}
where $U$ is a terminal value estimate from the learned value model.
MPPI iteratively refits $(\boldsymbol{\mu}, \boldsymbol{\sigma})$ to the highest-scoring (elite) candidates.
After the final iteration, $\boldsymbol{\mu}^*_{t:t+H}$ is not the weighted average of the elites but a single elite trajectory, sampled in proportion to its MPPI weight via the Gumbel-max trick~\citep{hansen2023tdmpc2}, while $\boldsymbol{\sigma}^*_{t:t+H}$ is the refitted standard deviation.
The planner outputs the optimized distribution $\mathcal{N}(\boldsymbol{\mu}_{t:t+H}^*, (\boldsymbol{\sigma}_{t:t+H}^*)^2)$, and only its first action is executed in the environment.
During training, this action is sampled from the optimized Gaussian distribution to explore the environment, whereas during evaluation the optimized mean $\boldsymbol{\mu}_t^*$ is executed.

\paragraph{Value functions.}
In the planning objective above, the terminal bootstrap term can be instantiated either as an action-value function $Q(\mathbf{z}_{t+H}, \mathbf{a}_{t+H})$ or as a state-value function $V(\mathbf{z}_{t+H})$.
Here, $Q^{p}(\mathbf{z}_t,\mathbf{a}_t) = \mathbb{E}_{p}[\sum_{k=0}^{\infty}\gamma^k r_{t+k} \mid \mathbf{z}_t,\mathbf{a}_t]$ is the expected return of taking action $\mathbf{a}_t$ at latent state $\mathbf{z}_t$ and following the policy prior $p$ thereafter; $V^{p}(\mathbf{z}_t) = \mathbb{E}_{\mathbf{a}_t \sim p(\cdot \mid \mathbf{z}_t)}[Q^{p}(\mathbf{z}_t,\mathbf{a}_t)]$ is the return under the policy prior.
The advantage $A^{p}(\mathbf{z}_t,\mathbf{a}_t) = Q^{p}(\mathbf{z}_t,\mathbf{a}_t) - V^{p}(\mathbf{z}_t)$ measures how much an action improves upon the return of the prior.
When only a state-value function is modeled explicitly, it can be trained via model-based TD learning~\citep{wang2025bootstrapped} by constructing an $N$-step bootstrapped target:
\begin{equation}
\mathcal{L}_V(\phi) = \mathbb{E}_{(s,a,r,s')} \left[\text{CE}\left(V_\phi(\mathbf{z}_t), \hat{V}(\mathbf{z}_t)\right)\right],
\quad
\hat{V}(\mathbf{z}_t) = \gamma^N V(\mathbf{z}_{t+N}) + \sum_{k=0}^{N-1} \gamma^k R(\mathbf{z}_{t+k}, \mathbf{a}_{t+k}),
\label{eq:model_based_td_value}
\end{equation}
where $\mathbf{z}_{t+1} = d(\mathbf{z}_t, \mathbf{a}_t)$ evolves through the learned model, $\hat{V}$ is the TD target computed from world-model rollouts, and $\text{CE}$ denotes cross-entropy with a discrete regression formulation of the value function.

\subsection{Diffusion Policies}
\label{subsec:ddpm}
A diffusion policy~\citep{ding2024diffusion,wang2024diffusion} models a conditional action distribution $p_\theta(\mathbf{a}\mid \mathbf{z})$ by transforming Gaussian noise into an action through an iterative denoising process conditioned on the latent state $\mathbf{z}$~\citep{sohl2015deep, ho2020ddpm}.
Given a clean action $\mathbf{a}_0$ and a diffusion step $\tau \in \{1,\dots,T\}$, a noisy action is constructed as $\mathbf{a}_\tau = \sqrt{\bar\alpha_\tau}\,\mathbf{a}_0 + \sqrt{1-\bar\alpha_\tau}\,\boldsymbol{\epsilon}$, where $\boldsymbol{\epsilon}\sim\mathcal{N}(\mathbf{0},\mathbf{I})$ and $\bar\alpha_\tau = \prod_{l=1}^{\tau}(1-\beta_l)$ for a predefined variance schedule $\beta_1,\dots,\beta_T$.
A denoising network $\boldsymbol{\epsilon}_\theta(\mathbf{a}_\tau,\tau,\mathbf{z})$ is trained to predict the injected noise by minimizing the denoising score-matching objective:
\begin{equation}
\mathbb{E}_{\mathbf{z},\mathbf{a}_0,\;\tau \sim \mathrm{Unif}\{1,\dots,T\},\;\boldsymbol{\epsilon}\sim\mathcal N(\mathbf{0},\mathbf{I})}
\left[
\left\|
\boldsymbol{\epsilon}-\boldsymbol{\epsilon}_\theta(\mathbf{a}_\tau,\tau,\mathbf{z})
\right\|^2
\right].
\label{eq:ddpm_loss}
\end{equation}
At inference time, an action is sampled by iteratively denoising for $T$ steps conditioned on $\mathbf{z}$.

\section{Methodology}
\label{sec:method}

\subsection{Value Estimation Error from Actor Divergence}
\label{sec:gap}

The planner relies on the value function to evaluate candidate trajectories: the terminal term $U$ in Eq.~\ref{eq:bmpc_planning_objective} accounts for the return beyond the planning horizon.
This value function, however, is learned under the policy prior, i.e., it estimates $V^{p}$, whereas the agent executes the planner.
It therefore predicts the planner's return accurately only when the prior and the planner have similar action distributions.
Any \emph{actor divergence} between the two biases the value estimate.
To analyze this divergence, we first characterize the action distribution of the planner.
At latent state $\mathbf{z}$, a single MPC call guided by the prior $p$ returns an optimized Gaussian distribution over action sequences.
Its first-action marginal, $\pi^\xi(\cdot\mid\mathbf{z},p)=\mathcal{N}\bigl(\boldsymbol{\mu}_0^{*,\xi}(\mathbf{z},p),\,(\boldsymbol{\sigma}_0^{*,\xi}(\mathbf{z},p))^2\bigr)$, depends on the internal randomness $\xi$ of the stochastic MPPI planner.
Marginalizing over $\xi$ yields the action distribution of the executed planner:
\begin{equation}
\pi(\cdot\mid\mathbf{z},p)=\mathbb{E}_{\xi}\bigl[\pi^{\xi}(\cdot\mid\mathbf{z},p)\bigr].
\label{eq:planner_distribution}
\end{equation}
Because MPPI optimizes a non-convex objective, independent draws of $\xi$ can converge to different local optima.
The planner distribution $\pi$ is therefore a mixture of Gaussians with distinct means, which is generally non-Gaussian.
To see how this affects the value estimate, we decompose its error with respect to the planner's return $V^{\pi}(\mathbf{z})$, i.e., the expected discounted return when executing $\pi$ from $\mathbf{z}$:
\begin{equation}
V_\phi(\mathbf{z}) - V^{\pi}(\mathbf{z}) = \underbrace{V_\phi(\mathbf{z}) - V^{p}(\mathbf{z})}_{\text{estimation error}} + \underbrace{V^{p}(\mathbf{z}) - V^{\pi}(\mathbf{z})}_{\text{actor-divergence bias}}.
\label{eq:value_error_decomposition}
\end{equation}
The first term is the error of $V_\phi$ in fitting its own target $V^{p}$, which is governed by model accuracy and function approximation.
The second term arises because the value is learned under the prior rather than the planner, and it vanishes when $p=\pi$.
An accurate value estimate thus requires the prior to match the planner distribution, i.e., $p\approx\pi$.

\begin{wraptable}{r}{0.42\textwidth}
\vspace{-12pt}
\centering
\caption{\textbf{Value estimation error on H1hand-Run.}
We evaluate the learned value network at three checkpoints during training,
reporting the relative mean absolute error (\%; lower is better), averaged over three seeds.}
\label{tab:value_error}
\vspace{-6pt}
\small
\setlength{\tabcolsep}{4pt}
\begin{tabular}{l*{3}{>{\centering\arraybackslash}p{3.2em}}}
\toprule
\multirow{2}{*}{Method} & \multicolumn{3}{c}{Environment steps (M)} \\
\cmidrule(lr){2-4}
& 0.4 & 1.2 & 2.0 \\
\midrule
BMPC & 82.4 & 54.6 & 75.9 \\
BOOM & 69.3 & 59.0 & 29.4 \\
\bottomrule
\end{tabular}
\vspace{-8pt}
\end{wraptable}

Recent methods reduce the actor divergence by aligning the prior with the planner.
BMPC~\citep{wang2025bootstrapped} minimizes the KL divergence from the planner distribution to the prior, while BOOM~\citep{zhan2025boom} maximizes the likelihood of planner actions under the prior.
Both methods, however, retain a Gaussian prior, which cannot represent the mixture in Eq.~\ref{eq:planner_distribution}.
The actor divergence thus persists no matter how well the Gaussian prior is trained.

Table~\ref{tab:value_error} reports the value error of BMPC and BOOM on H1hand-Run.
We measure the true return with a Monte Carlo estimate, following the protocol detailed in Appendix~\ref{app:value_error_protocol}.
Both Gaussian-prior methods retain a large value error throughout training.
The relative error of BMPC stays between 55\% and 82\%, while that of BOOM decreases from 69\% at 0.4M steps to 29\% at 2M steps.
Although the two methods train their priors and value functions differently, the error persists under both alignment objectives.
Removing this error therefore requires a policy prior expressive enough to represent the planner distribution.

\subsection{Diffusion Priors for MPC}
\label{subsec:dpmpc}

We propose \textbf{DPMPC} (Diffusion Priors for Model Predictive Control), a framework that trains a diffusion policy prior for guiding model-based planning.
DPMPC is built on the TD-MPC2~\citep{hansen2023tdmpc2} framework, inheriting the latent world model and MPPI planner.
We parameterize the policy $p_\theta(\mathbf{a} \mid \mathbf{z})$ as a diffusion model and use stochastic MPC to bootstrap the policy training.
DPMPC estimates the planner distribution with generalized lazy reanalyze, trains the diffusion policy prior on this estimate with advantage-weighted score matching, and learns the value function under the diffusion prior.
An overview of the DPMPC training loop is shown in Figure~\ref{fig:dpmpc_loop} and the complete training procedure is summarized in Algorithm~\ref{alg:dpmpc}.

\begin{algorithm}[t]
  \caption{DPMPC Training}
  \label{alg:dpmpc}
  \begin{algorithmic}[1]
  \Require $p_\theta$, $h$, $d$, $R$, $V_\phi$: randomly initialized networks
  \Statex \hspace*{\algorithmicindent} $\mathcal{B}$, $K$, $M$: replay buffer, reanalyze interval, queue capacity
  \While{not converged}
    \State \textcolor{gray}{\textit{// Collect experience:}}
    \For{step $t = 0 \ldots T_{\mathrm{env}}$}
      \State $\pi_t^{\xi} \leftarrow \pi^{\xi}(\cdot \mid h(\mathbf{s}_t),\, p_\theta)$; $\pi_{\text{lazy},t} \leftarrow \{\pi_t^{\xi}\}$ \Comment{Run MPC once; initialize queue}
      \State $\mathbf{a}_t \sim \pi_t^{\xi}$ \Comment{Sample and execute action}
      \State $(\mathbf{s}_{t+1},\, r_t) \leftarrow \mathrm{env.step}(\mathbf{a}_t)$ \Comment{Step environment}
      \State $\mathcal{B} \leftarrow \mathcal{B} \cup \bigl\{(\mathbf{s}_t,\, \mathbf{a}_t,\, r_t,\, \mathbf{s}_{t+1},\, \pi_{\text{lazy},t})\bigr\}$ \Comment{Add to buffer}
    \EndFor
    \State \textcolor{gray}{\textit{// Update networks:}}
    \For{num updates per episode}
      \State $\bigl(\mathbf{s}_t,\, \mathbf{a}_t,\, r_t,\, \mathbf{s}_{t+1},\, \pi_{\text{lazy},t}\bigr)_{t:t+H} \sim \mathcal{B}$ \Comment{Sample trajectory}
      \State Update $h$, $d$, $R$; update $V_\phi$ via Eq.~\ref{eq:model_based_td_value} \Comment{World model and value update}
      \State Update $p_\theta$ via Eq.~\ref{eq:final_diffusion_loss} \Comment{Diffusion policy prior update}
      \State \textcolor{gray}{\textit{// Generalized lazy reanalyze:}}
      \If{$\mathrm{update\_step} \bmod K = 0$}
        \State $\pi_{t:t+H}^{\xi} \leftarrow \pi^{\xi}(\cdot \mid h(\mathbf{s}_{t:t+H}),\, p_\theta)$ \Comment{Re-run MPC once per state}
        \State $\mathcal{B} \overset{\mathrm{enqueue}}{\longleftarrow} \pi_{t:t+H}^{\xi}$ \Comment{Append to FIFO queues; evict oldest if full}
      \EndIf
    \EndFor
  \EndWhile
  \end{algorithmic}
\end{algorithm}

\paragraph{Generalized lazy reanalyze.}
The training target for the diffusion policy prior is the planner distribution $\pi(\cdot\mid\mathbf{z}, p_\theta)$ in Eq.~\ref{eq:planner_distribution}.
Sampling directly from this distribution at every gradient step would require repeated MPC calls, incurring substantial planning cost.
We therefore store MPC realizations in the replay buffer for reuse across policy updates.
Since different stochastic MPC runs can reach different high-value action regions, we allow each replay state to retain up to $M$ realizations.
Concretely, for each replay state $\mathbf{s}$ with latent $\mathbf{z}$, the corresponding buffer slot maintains a first-in-first-out (FIFO) queue of $M_{\mathbf{z}} \le M$ recent MPC realizations $\bigl\{\pi^{\xi_i}(\cdot \mid \mathbf{z}, p_{\theta_i})\bigr\}_{i=1}^{M_{\mathbf{z}}}$.
During environment interaction, the MPC call at $\mathbf{z}_t$ initializes the queue with a single realization ($M_{\mathbf{z}} = 1$).
At fixed update intervals, we rerun MPC once for each state in a sampled batch using fresh planner randomness, append the new realization to its queue, and evict the oldest realization if the queue is full.
The stored realizations define the empirical estimate of the planner distribution through their uniform mixture:
$
\pi_{\text{lazy}}(\cdot \mid \mathbf{z})
:=
\frac{1}{M_{\mathbf{z}}}
\sum_{i=1}^{M_{\mathbf{z}}}
\pi^{\xi_i}(\cdot \mid \mathbf{z}, p_{\theta_i}).
$
This procedure extends the \emph{lazy reanalyze} mechanism of BMPC to support multiple MPC realizations per state.

\paragraph{Advantage-weighted score matching.}
Some MPC realizations can be stale since the buffer stores multiple realizations per state and the refresh rate is limited.
In policy training, we therefore seek a training objective that emphasizes actions with higher estimated value while keeping the prior close to the planner distribution, so that the value learned under the prior remains accurate for the planner.
By the definition of advantage in Section~\ref{subsec:planning_based_mbrl}, emphasizing higher-value actions is equivalent to emphasizing higher-advantage actions.
Let $d_{\mathcal{B}}(\mathbf{z})$ denote the latent-state distribution induced by $\mathcal{B}$.
We formulate policy training as a constrained optimization problem:
\begin{equation}
\label{eq:awr_kl}
\begin{aligned}
\arg\max_{p_{\theta}} \quad & \int_{\mathbf{z}} d_{\mathcal{B}}(\mathbf{z}) \int_{\mathbf{a}} p_{\theta}(\mathbf{a}|\mathbf{z}) \, A(\mathbf{z},\mathbf{a}) \; d\mathbf{a} \, d\mathbf{z} \\
\text{s.t.} \quad & D_{\mathrm{KL}}\!\left(p_{\theta}(\cdot|\mathbf{z}) \,\Big\|\, \pi_{\text{lazy}}(\cdot|\mathbf{z})\right) \leq \epsilon, \quad \forall\, \mathbf{z}.
\end{aligned}
\end{equation}
Solving this problem in closed form and projecting the solution onto the diffusion policy prior yields an advantage-weighted score-matching objective.
The full derivation is deferred to Appendix~\ref{app:awr_derivation}.
We train the diffusion policy prior with this objective on $H$-step latent rollouts of the learned world model and apply temporal weighting $\lambda^t$ along the rollout horizon, which yields the final DPMPC loss
\begin{equation}
\begin{gathered}
\begin{aligned}
\mathcal{L}_{\mathrm{DPMPC}}(\theta)
&=
\mathbb{E}_{(\mathbf{s}, \mathbf{a}, \pi_{\text{lazy}})_{0:H} \sim \mathcal{B}}
\Biggl[
\sum_{t=0}^{H} \lambda^t \\
&\hspace{-10pt}\cdot\;
\mathbb{E}_{\substack{\tau \sim \mathrm{Unif}\{1,\dots,T\},\; \boldsymbol{\epsilon} \sim \mathcal{N}(\mathbf{0}, \mathbf{I})\\ \mathbf{a}_t^{\text{lazy}} \sim \pi_{\text{lazy},t}(\cdot|\mathbf{z}_t)}}
\left[
w(\mathbf{z}_t, \mathbf{a}_t^{\text{lazy}})
\left\|
\boldsymbol{\epsilon} - \boldsymbol{\epsilon}_\theta\!\left(
\sqrt{\bar{\alpha}_\tau}\, \mathbf{a}_t^{\text{lazy}} + \sqrt{1 - \bar{\alpha}_\tau}\, \boldsymbol{\epsilon},\;
\tau,\; \mathbf{z}_t
\right)
\right\|^2
\right]
\Biggr],
\end{aligned} \\
w(\mathbf{z}_t, \mathbf{a}_t^{\text{lazy}}) = \exp\!\left(\frac{\hat{A}(\mathbf{z}_t, \mathbf{a}_t^{\text{lazy}})}{\eta\, S}\right),
\end{gathered}
\label{eq:final_diffusion_loss}
\end{equation}
where $\mathbf{z}_0 = h(\mathbf{s}_0)$ and $\mathbf{z}_{t+1} = d(\mathbf{z}_t, \mathbf{a}_t)$ are rolled out on the world model.
Here, $w(\mathbf{z}_t, \mathbf{a}_t^{\text{lazy}})$ is the per-sample advantage weight and $\eta > 0$ is the advantage temperature.
Small $\eta$ sharpens the weighting toward high-advantage actions; in the limit $\eta \to +\infty$, all weights collapse to a constant and the objective reduces to imitation of $\pi_{\text{lazy}}$.
We normalize the advantage by a running scale $S$, a moving estimate of the spread of $\hat{A}$ over training batches, so that a single temperature $\eta$ transfers across tasks with different reward scales.
The weights are also clamped to a finite range for training stability.
The advantage is estimated from the learned world model via the one-step bootstrap $\hat{A}(\mathbf{z},\mathbf{a}) = R(\mathbf{z},\mathbf{a}) + \gamma V_\phi(d(\mathbf{z},\mathbf{a})) - V_\phi(\mathbf{z})$, where $V_\phi$ is the learned latent state-value function.

\paragraph{Value function training.}
DPMPC trains $V_\phi$ with the model-based TD objective in Eq.~\ref{eq:model_based_td_value}, sampling every rollout action from the diffusion policy prior.
The value function $V_\phi$ therefore approximates $V^{p_\theta}$.
Since the diffusion prior is trained to match the planner distribution, $V^{p_\theta}$ provides an estimate of the value of the planner.

\section{Experiments}
\label{sec:experiments}

\subsection{Experimental Setup}

We compare DPMPC with six representative online RL baselines, including two model-free methods and four model-based methods. For the model-free baselines, we choose \textbf{SAC}~\citep{haarnoja2018sac}, which is built on a Gaussian policy under the maximum-entropy RL framework, and \textbf{SDAC}~\citep{ma2025efficient}, a diffusion-policy-based RL method. For the model-based methods, \textbf{DreamerV3}~\citep{hafner2025mastering} represents imagination-based MBRL, while \textbf{TD-MPC2}~\citep{hansen2023tdmpc2}, \textbf{BMPC}~\citep{wang2025bootstrapped}, and \textbf{BOOM}~\citep{zhan2025boom} are planning-driven approaches that differ in how they train the Gaussian policy prior. Experiments are conducted on 14 challenging high-dimensional tasks collected from DMControl~\citep{tassa2018dmcontrol}, MyoSuite~\citep{MyoSuite2022}, and HumanoidBench~\citep{sferrazza2024humanoidbench}. Complete environment specifications and hyperparameter settings are provided in Appendix~\ref{app:exp_details}.

\begin{figure}[t]
\centering
\includegraphics[width=\textwidth]{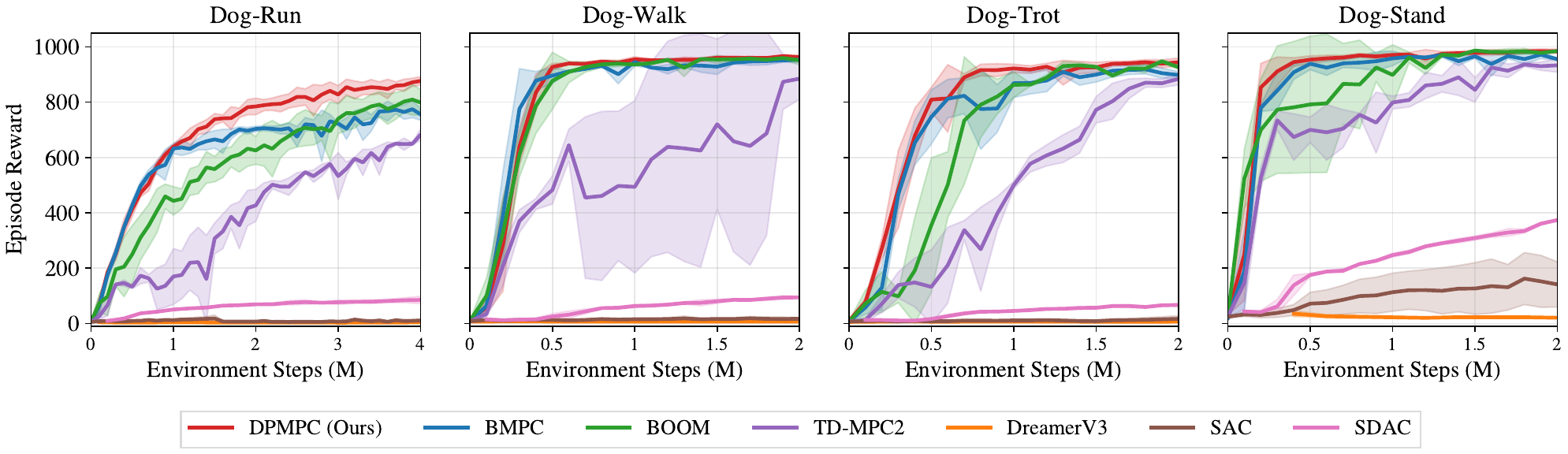}
\caption{\textbf{Training curves on DMControl.}
Dog-Run is trained for 4M environment steps due to its slower convergence, while all other tasks are trained for 2M steps.}
\label{fig:dmcontrol_dog}
\end{figure}

\begin{figure}[t]
\centering
\includegraphics[width=0.75\textwidth]{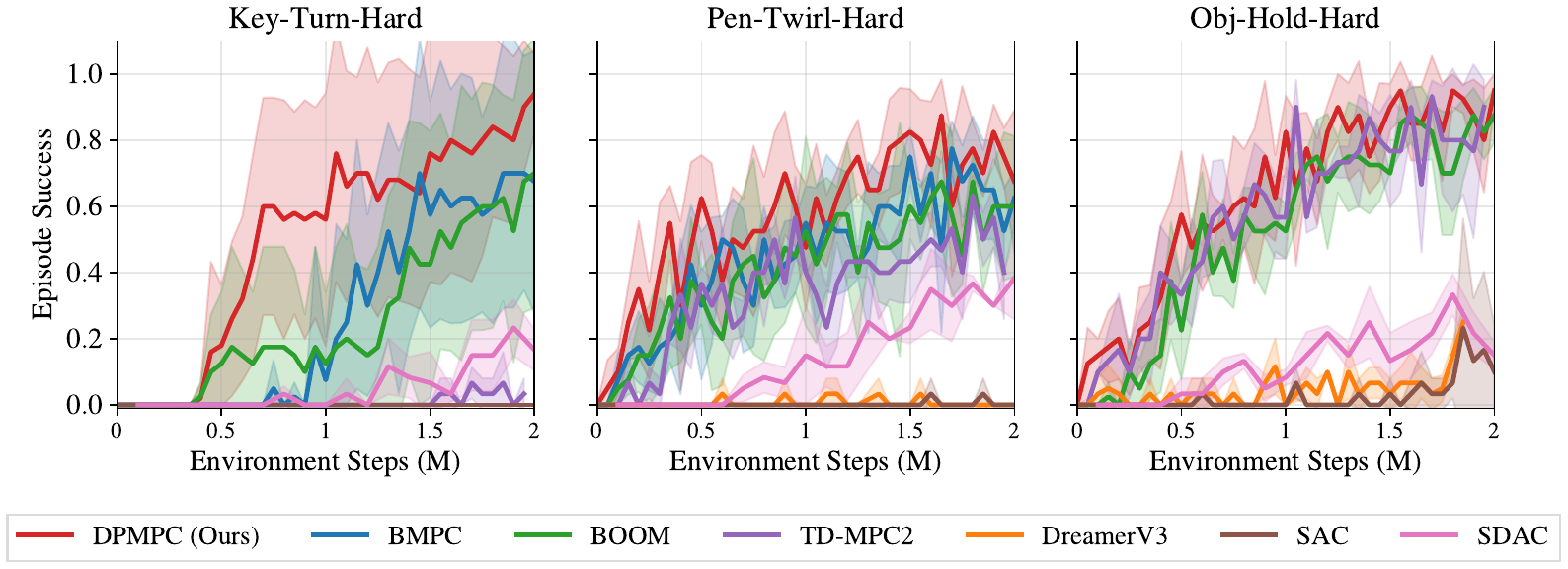}
\caption{\textbf{Training curves on MyoSuite.}
Performance on the dexterous manipulation tasks is reported as success rate.}
\label{fig:myosuite}
\end{figure}

\subsection{Comparative Evaluation}
\label{subsec:comparative}

\begin{figure}[t]
\centering
\includegraphics[width=\textwidth]{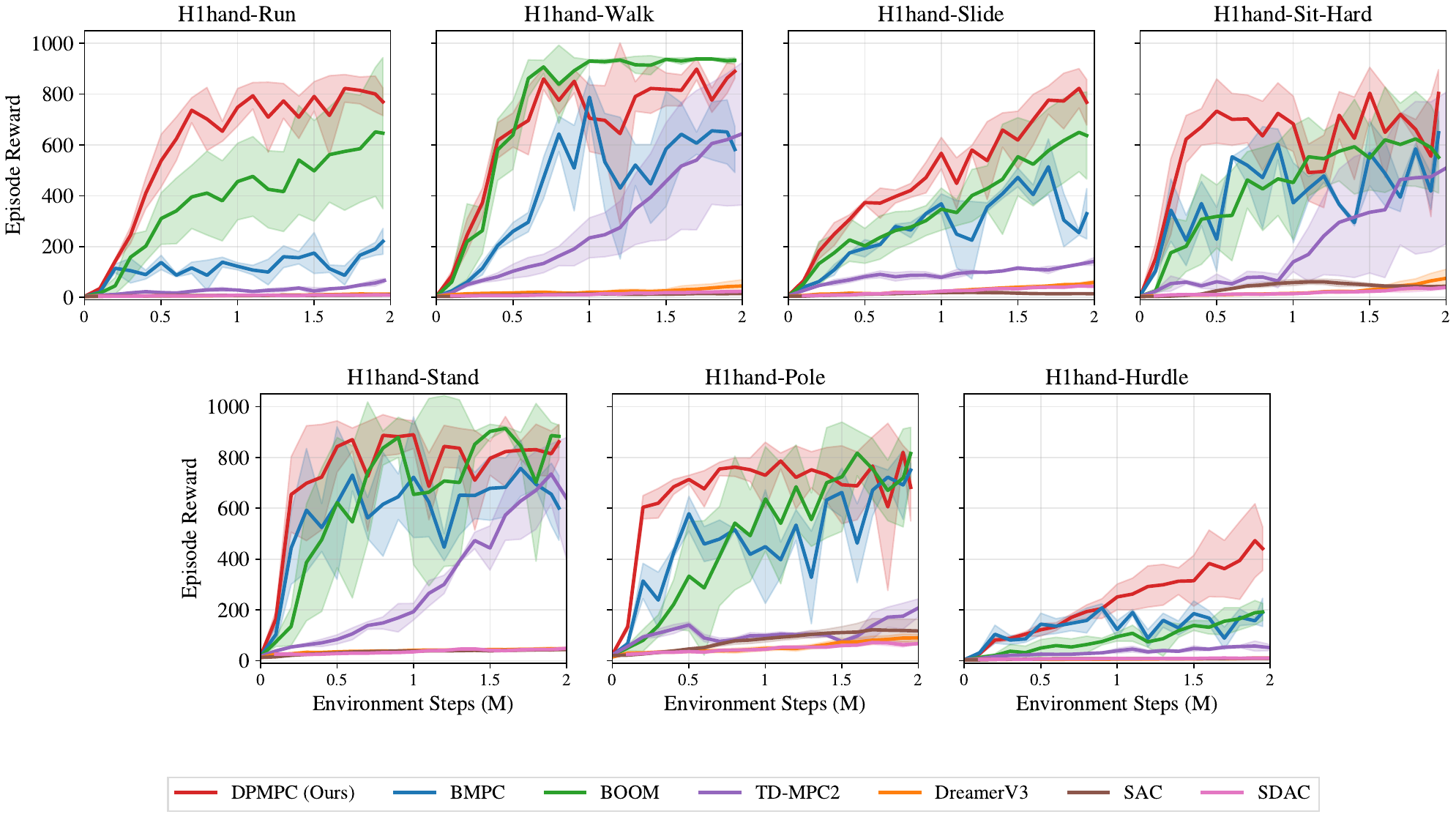}
\caption{\textbf{Training curves on HumanoidBench.}
The H1hand tasks form a challenging high-dimensional control setting with a 61-dimensional action space.
The top-left panel reports the average episode return across all seven tasks.}
\label{fig:h1hand}
\end{figure}

We report training curves across three benchmarks in Figures~\ref{fig:dmcontrol_dog}--\ref{fig:h1hand}, averaged over four seeds.
DPMPC consistently outperforms the model-free baselines (SAC, SDAC) and the imagination-based DreamerV3 across all benchmarks.
Among the planning-driven baselines, DPMPC achieves the highest asymptotic return and sample efficiency on DMControl and MyoSuite.
On the 61-dimensional H1hand tasks, DPMPC outperforms every baseline with an average return of \textbf{745}.
This marks a \textbf{+12.0\%} improvement over BOOM, the strongest competing planner, and a \textbf{+57.1\%} gain over BMPC.
The margin widens to \textbf{+130.9\%} over TD-MPC2, suggesting that the diffusion policy prior is especially effective for planning in high-dimensional action spaces.
These gains come at a moderate computational cost: DPMPC requires about $1.3\times$ the wall-clock training time of BMPC and less than half that of BOOM (Table~\ref{tab:training_time}; see Appendix~\ref{app:training_time} for the hardware setup).

\begin{table}[t]
\centering
\caption{\textbf{Training time.} Wall-clock hours for 2M environment steps on a single RTX 3090 GPU.}
\label{tab:training_time}
\small
\begin{tabular}{lcccc}
\toprule
Task & TD-MPC2 & BMPC & BOOM & DPMPC (ours) \\
\midrule
Dog-Run & 5.3 & 7.5 & 20.1 & 9.6 \\
H1hand-Run & 11.0 & 15.6 & 40.4 & 19.3 \\
\bottomrule
\end{tabular}
\end{table}

\subsection{Value Estimation Analysis}
\label{subsec:value_analysis}

Table~\ref{tab:value_analysis} repeats the diagnostic of Section~\ref{sec:gap} for all three methods on Dog-Run and H1hand-Run.
On H1hand-Run, DPMPC attains the lowest value error at every checkpoint.
Its relative error drops from 51\% at 0.4M steps to 6.5\% at 1.2M steps and stays below 7\% thereafter.
BMPC and BOOM retain errors of at least 29\% throughout training.
On Dog-Run, BMPC is more accurate at 0.8M steps, but DPMPC reduces its error to 2\% at 2.4M steps and 1\% at 4M steps, whereas BMPC and BOOM plateau at about 5\% and 10\%, respectively.
The advantage of DPMPC in value accuracy is thus largest on the task with the higher-dimensional action space (61 dimensions in H1hand-Run versus 38 in Dog-Run), which is also where DPMPC achieves its largest performance gains.
We attribute the remaining error of DPMPC on both tasks partly to generalized lazy reanalyze.
Because the planner keeps evolving during training, the stored MPC realizations lag behind the current planner, so $\pi_{\text{lazy}}$ tracks the planner distribution only approximately.
Appendix~\ref{app:additional_viz} further shows that the learned diffusion prior produces non-Gaussian, often multimodal action distributions.

\begin{table}[t]
\centering
\caption{\textbf{Value estimation analysis.}
Relative mean absolute error (\%) of the learned value at three training checkpoints (lower is better).
Entries report the mean $\pm$ standard deviation over three seeds.
Bold marks the lowest mean error in each column.}
\label{tab:value_analysis}
\small
\setlength{\tabcolsep}{4pt}
\begin{tabular}{lcccccc}
\toprule
 & \multicolumn{3}{c}{Dog-Run} & \multicolumn{3}{c}{H1hand-Run} \\
\cmidrule(lr){2-4} \cmidrule(lr){5-7}
Method & 0.8M & 2.4M & 4M & 0.4M & 1.2M & 2M \\
\midrule
BMPC & \textbf{7.1}{\scriptsize$\pm$5.7} & 5.5{\scriptsize$\pm$0.7} & 5.4{\scriptsize$\pm$3.7} & 82.4{\scriptsize$\pm$34.1} & 54.6{\scriptsize$\pm$21.4} & 75.9{\scriptsize$\pm$35.3} \\
BOOM & 12.0{\scriptsize$\pm$5.5} & 11.1{\scriptsize$\pm$3.4} & 9.5{\scriptsize$\pm$1.7} & 69.3{\scriptsize$\pm$3.3} & 59.0{\scriptsize$\pm$13.9} & 29.4{\scriptsize$\pm$1.8} \\
DPMPC (ours) & 12.4{\scriptsize$\pm$2.7} & \textbf{1.8}{\scriptsize$\pm$1.1} & \textbf{1.0}{\scriptsize$\pm$0.6} & \textbf{50.6}{\scriptsize$\pm$6.1} & \textbf{6.5}{\scriptsize$\pm$2.6} & \textbf{6.8}{\scriptsize$\pm$5.6} \\
\bottomrule
\end{tabular}
\end{table}

\begin{figure}[t]
\centering
\begin{subfigure}[b]{0.24\textwidth}
    \centering
    \includegraphics[width=\textwidth]{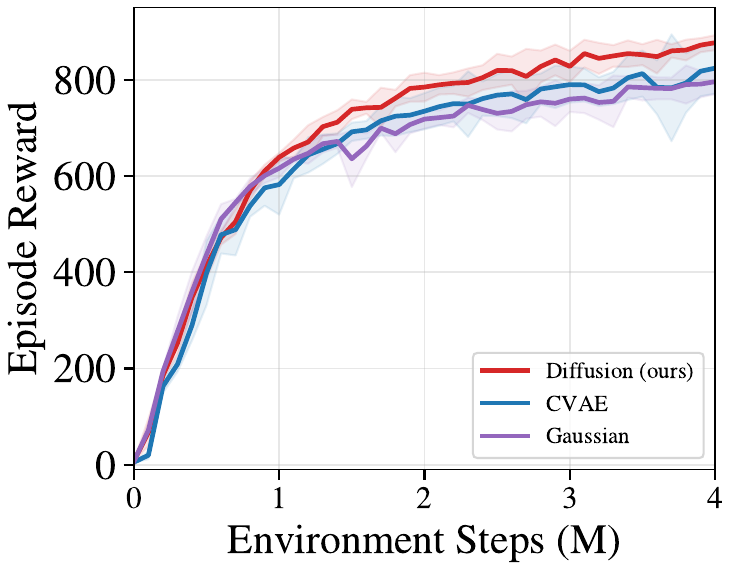}
    \caption{}
    \label{fig:ablation_policy}
\end{subfigure}
\hfill
\begin{subfigure}[b]{0.24\textwidth}
    \centering
    \includegraphics[width=\textwidth]{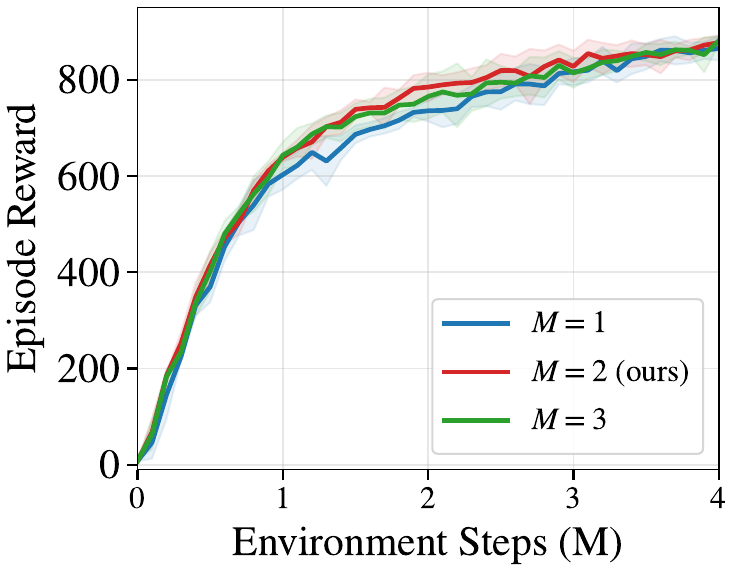}
    \caption{}
    \label{fig:ablation_M}
\end{subfigure}
\hfill
\begin{subfigure}[b]{0.24\textwidth}
    \centering
    \includegraphics[width=\textwidth]{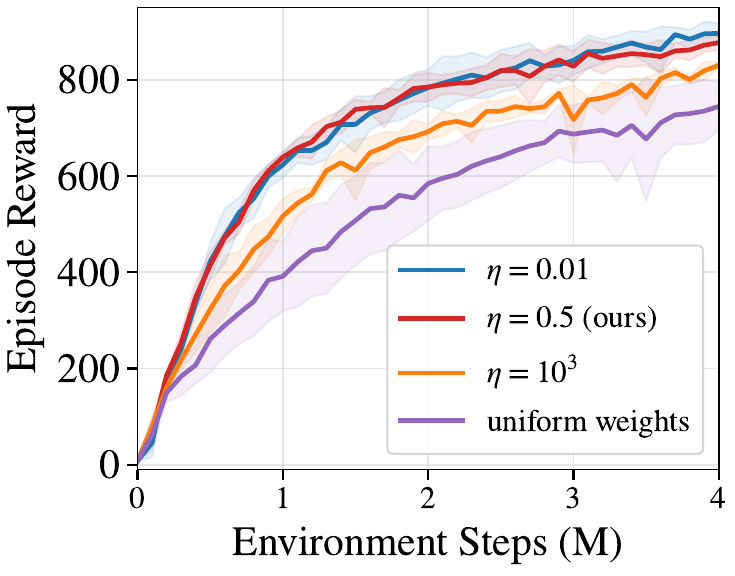}
    \caption{}
    \label{fig:ablation_eta}
\end{subfigure}
\hfill
\begin{subfigure}[b]{0.24\textwidth}
    \centering
    \includegraphics[width=\textwidth]{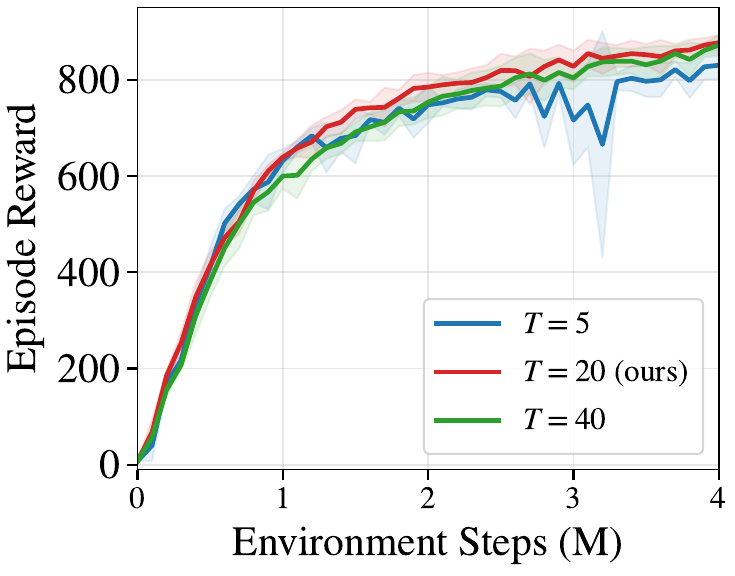}
    \caption{}
    \label{fig:ablation_steps}
\end{subfigure}
\caption{\textbf{Ablation studies on Dog-Run.}
(\subref{fig:ablation_policy}) Policy prior parameterization.
(\subref{fig:ablation_M}) Number of MPC realizations stored per state $M$.
(\subref{fig:ablation_eta}) Advantage temperature $\eta$.
(\subref{fig:ablation_steps}) Number of diffusion steps $T$.}
\label{fig:ablation_dog}
\end{figure}

\subsection{Ablation Studies}

\paragraph{Effect of the policy parameterization.}
Figure~\ref{fig:ablation_policy} compares three parameterizations of the policy prior: the diffusion policy of DPMPC, a conditional variational autoencoder (CVAE)~\citep{diederik2019introduction}, and a Gaussian policy.
The CVAE and the Gaussian policy each replace the diffusion policy within DPMPC, leaving all other components unchanged.
The CVAE and the Gaussian policy perform comparably, and the diffusion policy achieves the highest return.
This result shows that the expressiveness of the policy prior is important for control performance.
The gap widens substantially on H1hand-Run, where both alternatives fall well behind the diffusion policy (Appendix~\ref{app:additional_ablations}).

\paragraph{Effect of the number of MPC realizations.}
Figure~\ref{fig:ablation_M} varies the number of MPC realizations $M \in \{1, 2, 3\}$ stored per buffer state during generalized lazy reanalyze.
Setting $M=2$ yields the best sample efficiency and shows stable training.
With $M=1$, each target reflects only a single MPC realization.
This choice slows learning, likely because the single realization fails to capture the planner distribution.
Increasing $M$ to $3$ brings no further gain, possibly because older realizations become stale.
Appendix~\ref{app:additional_ablations} further isolates the periodic refresh of generalized lazy reanalyze.

\paragraph{Effect of advantage temperature.}
Figure~\ref{fig:ablation_eta} examines the advantage temperature $\eta$ alongside an unweighted variant that assigns uniform weights to all targets.
Removing the weighting is the most damaging setting: it learns most slowly throughout training and attains the lowest asymptotic return.
With $\eta=10^3$ the weights become nearly uniform and performance degrades toward that variant, confirming that the gains stem from advantage weighting rather than from score matching on $\pi_{\text{lazy}}$ alone.
Sharpening the weighting from $\eta=0.5$ to $\eta=0.01$ leaves performance essentially unchanged, indicating that DPMPC is robust to $\eta$ over this range.

\paragraph{Effect of the number of diffusion steps.}
Figure~\ref{fig:ablation_steps} varies the number of denoising steps of the diffusion policy prior.
With $T=5$, the short denoising chain limits the expressiveness of the prior, which leads to unstable learning after 2.5M steps and a lower final return.
Increasing $T$ to $40$ slightly slows learning in the intermediate phase without improving the final return, while doubling the cost of sampling from the prior.

\section{Related Work}
\paragraph{Model-based Reinforcement Learning.}
Model-based reinforcement learning improves sample efficiency by learning a dynamics model and using it to support decision-making.
Imagination-based methods, such as Dreamer and DIAMOND~\citep{hafner2025mastering, alonso2024diffusion}, optimize policies with synthetic rollouts generated by the learned model and have achieved strong results in visual control.
Planning-driven methods such as the TD-MPC family, BMPC, BOOM, and TD-M(PC)$^2$ combine learned latent models with MPC and use a learned policy prior to guide trajectory optimization~\citep{Hansen2022tdmpc,hansen2023tdmpc2,wang2025bootstrapped,zhan2025boom, lin2025td}.
While these methods achieve strong sample efficiency, they learn the value function under the policy prior but execute the planner, and this \emph{actor divergence}~\citep{sikchi2021learning,zhan2025boom} biases the value estimate.
Their policy priors are typically Gaussian, which cannot represent the non-Gaussian action distribution of the planner and thus leaves a residual actor divergence.
The closest prior work is BMPC, which narrows this divergence by imitating the planner distribution with a Gaussian prior, yet a residual divergence remains.
Our approach, DPMPC, addresses this limitation by using a diffusion policy prior that can capture the multimodal planner distribution and reduce the actor divergence.

\paragraph{Diffusion Policies for Online RL.}
Diffusion policies have recently been adopted as expressive actors for online reinforcement learning, where their iterative denoising process can represent complex action distributions beyond standard Gaussian policies~\citep{wang2024diffusion,ding2024diffusion,psenka2024learning,ma2025efficient,dong2025maximum}. Existing methods are largely model-free and couple diffusion policy optimization to a learned critic. One line of work maximizes value estimates by backpropagating through the denoising chain~\citep{wang2024diffusion}, but this direct approach is computationally expensive and can be unstable because gradients must pass through many sequential sampling steps. 
Another line avoids full backpropagation by using critic-guided objectives, including Q-score matching~\citep{psenka2024learning} and Q-weighted variational objectives~\citep{ding2024diffusion}. 
Although effective, these methods define target action distributions only implicitly through the learned value function, so policy updates are typically formed from samples drawn from the current policy rather than from explicit high-value action targets, which can make diffusion policy learning less direct and less sample efficient~\citep{ma2025efficient}. In contrast, DPMPC uses model-based planning to produce explicit high-value action targets for training the diffusion policy prior.
Closer to our setting, recent works combine diffusion policies with learned world models and planning~\citep{balim2026modelbasedonline,asri2026modip}.
GTP~\citep{balim2026modelbasedonline} also trains a diffusion policy on planner-generated targets weighted by value estimates, but its ensemble world model and beam search with iterative denoising incur substantially higher training cost per update.
MODIP~\citep{asri2026modip} uses MPC as a policy-improvement operator for a diffusion policy pretrained on offline data via behavior cloning, targeting offline-to-online fine-tuning, whereas DPMPC learns entirely online from scratch.

\paragraph{Advantage-Weighted Regression Methods.}
DPMPC is also related to advantage-weighted regression methods such as AWR~\citep{peng2019awr} and AWAC~\citep{nair2020awac}, which update a policy by fitting actions from a replay buffer with weights determined by their estimated advantages. This connection is mainly at the level of the policy-improvement objective: both AWAC and DPMPC can be viewed as projecting an advantage-weighted target distribution onto a parameterized policy. However, the source and role of the supervised targets are different. AWAC learns from actions contained in an offline dataset or replay buffer, whereas DPMPC trains a diffusion prior from planner-generated action distributions that are stored in the replay buffer and periodically refreshed through generalized lazy reanalyze. 
Thus, DPMPC does not use advantage weighting for offline policy extraction; instead, it uses it to prioritize higher-value planner targets during policy training.

\section{Conclusion}

We introduce DPMPC, a planning-driven MBRL framework in which a diffusion policy serves as an expressive action prior for MPC.
Trained on the planner distribution estimated from generalized lazy reanalyze, the diffusion prior reduces the actor divergence, so that the value function learned under the prior provides a more accurate estimate of the planner's return.
Across DMControl, MyoSuite, and HumanoidBench, DPMPC achieves state-of-the-art asymptotic performance and improved sample efficiency, with the most pronounced gains on HumanoidBench.
Value estimation analysis shows that DPMPC achieves the lowest value error.
These results establish policy-prior parameterization as an important design factor in planning-driven MBRL.
Future work may explore other expressive policy representations.
Faster generative models, such as consistency models and flow-matching models, could reduce the training and inference cost while maintaining performance.
These models might also be used as expressive priors to speed up the training of our framework.

\bibliography{ref}
\bibliographystyle{iclr2027_conference}

\newpage
\appendix
\section{Derivation of the Advantage-Weighted Denoising Score-Matching Objective}
\label{app:awr_derivation}

We derive the advantage-weighted score-matching objective underlying Eq.~\ref{eq:final_diffusion_loss} from the constrained policy optimization problem in Eq.~\ref{eq:awr_kl}. We first solve the KL-regularized optimization problem in closed form, which yields an optimal but intractable target policy. We then project this target onto the parameterized diffusion policy and replace the intractable log-likelihood of the diffusion model with its variational lower bound, which leads to the advantage-weighted score-matching objective.

\paragraph{Problem restatement.}
Let $\pi_{\text{lazy}}(\cdot|\mathbf{z})$ denote the lazily refreshed MPC target distribution stored in the replay buffer $\mathcal{B}$, let $d_{\mathcal{B}}(\mathbf{z})$ denote the latent-state distribution induced by $\mathcal{B}$, and let $A(\mathbf{z},\mathbf{a})$ denote the advantage defined in Section~\ref{subsec:planning_based_mbrl}. Recalling Eq.~\ref{eq:awr_kl}, we seek a diffusion policy $p_{\theta}(\mathbf{a}|\mathbf{z})$ that maximizes the expected advantage while staying close to $\pi_{\text{lazy}}$ at every latent state:
\begin{equation}
\arg\max_{p_{\theta}} \;\; \int_{\mathbf{z}} d_{\mathcal{B}}(\mathbf{z}) \int_{\mathbf{a}} p_{\theta}(\mathbf{a}|\mathbf{z})\, A(\mathbf{z},\mathbf{a}) \, d\mathbf{a}\, d\mathbf{z},
\quad
\text{s.t.}\;\; D_{\mathrm{KL}}\!\left(p_{\theta}(\cdot|\mathbf{z}) \,\|\, \pi_{\text{lazy}}(\cdot|\mathbf{z})\right) \leq \epsilon, \;\; \forall\, \mathbf{z}.
\label{eq:app_problem_restated}
\end{equation}

\paragraph{Optimal policy under KL regularization.}
Enforcing the pointwise KL constraint in Eq.~\ref{eq:app_problem_restated} at every latent state would require a separate Lagrange multiplier per state, which is intractable. Following \citet{peng2019awr}, we therefore relax it to a single KL bound in expectation over the replay-buffer state distribution,
\begin{equation}
\int_{\mathbf{z}} d_{\mathcal{B}}(\mathbf{z})\, D_{\mathrm{KL}}\!\left(p(\cdot|\mathbf{z}) \,\|\, \pi_{\text{lazy}}(\cdot|\mathbf{z})\right) d\mathbf{z} \;\leq\; \epsilon,
\end{equation}
which is controlled by a single temperature rather than a per-state one. Introducing a Lagrange multiplier $\eta$ ($\eta > 0$) for this soft KL constraint and per-state multipliers $\alpha_{\mathbf{z}}$ for the normalization constraint $\int_{\mathbf{a}} p(\mathbf{a}|\mathbf{z})\, d\mathbf{a} = 1$ yields the Lagrangian~\citep{peng2019awr,peters2010reps}
\begin{equation}
\begin{aligned}
\mathcal{L}(p,\eta,\alpha)
&= \int_{\mathbf{z}} d_{\mathcal{B}}(\mathbf{z}) \int_{\mathbf{a}} p(\mathbf{a}|\mathbf{z})\, A(\mathbf{z},\mathbf{a})\, d\mathbf{a}\, d\mathbf{z} \\
&\quad +\, \eta\!\left( \epsilon - \int_{\mathbf{z}} d_{\mathcal{B}}(\mathbf{z})\, D_{\mathrm{KL}}\!\left(p(\cdot|\mathbf{z}) \,\|\, \pi_{\text{lazy}}(\cdot|\mathbf{z})\right) d\mathbf{z} \right) \\
&\quad +\, \int_{\mathbf{z}} \alpha_{\mathbf{z}}\!\left( 1 - \int_{\mathbf{a}} p(\mathbf{a}|\mathbf{z})\, d\mathbf{a} \right) d\mathbf{z}.
\end{aligned}
\label{eq:app_lagrangian}
\end{equation}
Taking the functional derivative with respect to $p(\mathbf{a}|\mathbf{z})$ and setting it to zero gives
\begin{equation}
\frac{\delta \mathcal{L}}{\delta p(\mathbf{a}|\mathbf{z})} = d_{\mathcal{B}}(\mathbf{z})\, A(\mathbf{z},\mathbf{a}) - \eta\, d_{\mathcal{B}}(\mathbf{z})\!\left( \log \frac{p(\mathbf{a}|\mathbf{z})}{\pi_{\text{lazy}}(\mathbf{a}|\mathbf{z})} + 1 \right) - \alpha_{\mathbf{z}} = 0.
\end{equation}
Dividing through by $\eta\, d_{\mathcal{B}}(\mathbf{z})$ and rearranging yields
\begin{equation}
\log p(\mathbf{a}|\mathbf{z}) = \log \pi_{\text{lazy}}(\mathbf{a}|\mathbf{z}) + \frac{A(\mathbf{z},\mathbf{a})}{\eta} - 1 - \frac{\alpha_{\mathbf{z}}}{\eta\, d_{\mathcal{B}}(\mathbf{z})}.
\end{equation}
Exponentiating both sides gives
\begin{equation}
p(\mathbf{a}|\mathbf{z}) = \pi_{\text{lazy}}(\mathbf{a}|\mathbf{z})\, \exp\!\left(\frac{A(\mathbf{z},\mathbf{a})}{\eta}\right) \exp\!\left( -\frac{\alpha_{\mathbf{z}}}{\eta\, d_{\mathcal{B}}(\mathbf{z})} - 1 \right).
\label{eq:app_unnormalized_policy}
\end{equation}
The second exponential factor depends on $\mathbf{z}$ but is independent of $\mathbf{a}$. Therefore, for a fixed latent state, it is constant across all actions and can be interpreted as a $\mathbf{z}$-dependent normalizing factor $1/Z(\mathbf{z})$. Enforcing the normalization condition $\int_{\mathbf{a}} p(\mathbf{a}|\mathbf{z})\, d\mathbf{a} = 1$ in Eq.~\ref{eq:app_unnormalized_policy} determines its value:
\begin{equation}
Z(\mathbf{z}) = \exp\!\left( \frac{\alpha_{\mathbf{z}}}{\eta\, d_{\mathcal{B}}(\mathbf{z})} + 1 \right) = \int_{\mathbf{a}'} \pi_{\text{lazy}}(\mathbf{a}'|\mathbf{z})\, \exp\!\left( \frac{A(\mathbf{z},\mathbf{a}')}{\eta} \right) d\mathbf{a}'.
\label{eq:app_partition}
\end{equation}
The optimal policy is therefore
\begin{equation}
p^*(\mathbf{a}|\mathbf{z}) = \frac{1}{Z(\mathbf{z})}\, \pi_{\text{lazy}}(\mathbf{a}|\mathbf{z})\, \exp\!\left(\frac{A(\mathbf{z},\mathbf{a})}{\eta}\right).
\label{eq:app_optimal_policy}
\end{equation}

\paragraph{Advantage-weighted score-matching objective.}
The optimal policy $p^*$ in Eq.~\ref{eq:app_optimal_policy} is defined only implicitly through the advantage function and cannot be directly represented by the diffusion model. We therefore project $p^*$ onto the parameterized diffusion policy $p_\theta$ by minimizing the forward KL divergence~\citep{peng2019awr}, weighted by the same state distribution $d_{\mathcal{B}}$ inherited from Eq.~\ref{eq:app_problem_restated}:
\begin{equation}
\min_{\theta} \;\; \mathbb{E}_{\mathbf{z} \sim d_{\mathcal{B}}}\!\left[ D_{\mathrm{KL}}\!\left(p^*(\cdot|\mathbf{z}) \,\|\, p_\theta(\cdot|\mathbf{z})\right) \right].
\label{eq:app_kl_projection}
\end{equation}
Expanding the KL divergence gives
\begin{equation}
\min_{\theta} \;\; \mathbb{E}_{\mathbf{z} \sim d_{\mathcal{B}}}\!\left[ \mathbb{E}_{\mathbf{a} \sim p^*(\cdot|\mathbf{z})}\!\left[\log p^*(\mathbf{a}|\mathbf{z}) - \log p_\theta(\mathbf{a}|\mathbf{z})\right] \right].
\end{equation}
The first term is independent of $\theta$, so the problem is equivalently written as
\begin{equation}
\max_{\theta} \;\; \mathbb{E}_{\mathbf{z} \sim d_{\mathcal{B}}}\!\left[ \mathbb{E}_{\mathbf{a} \sim p^*(\cdot|\mathbf{z})}\!\left[ \log p_\theta(\mathbf{a}|\mathbf{z}) \right] \right].
\end{equation}
To avoid sampling from the intractable policy $p^*$, we rewrite the inner expectation using $\pi_{\text{lazy}}$ as the proposal distribution, while the outer expectation remains over the replay-buffer state distribution $\mathbf{z} \sim \mathcal{B}$:
\begin{equation}
\max_{\theta} \;\; \mathbb{E}_{\mathbf{z} \sim \mathcal{B},\; \mathbf{a}_0 \sim \pi_{\text{lazy}}(\cdot|\mathbf{z})}\!\left[ \frac{p^*(\mathbf{a}_0|\mathbf{z})}{\pi_{\text{lazy}}(\mathbf{a}_0|\mathbf{z})} \log p_\theta(\mathbf{a}_0|\mathbf{z}) \right].
\end{equation}
Substituting Eq.~\ref{eq:app_optimal_policy} into the above equation yields
\begin{equation}
\max_{\theta} \;\; \mathbb{E}_{\mathbf{z} \sim \mathcal{B},\; \mathbf{a}_0 \sim \pi_{\text{lazy}}(\cdot|\mathbf{z})}\!\left[ \frac{1}{Z(\mathbf{z})}\exp\!\left(\frac{A(\mathbf{z},\mathbf{a}_0)}{\eta}\right) \log p_\theta(\mathbf{a}_0|\mathbf{z}) \right].
\end{equation}
Since $Z(\mathbf{z}) > 0$ does not depend on $\theta$, it is dropped from the optimization for simplicity~\citep{peng2019awr}, giving the advantage-weighted maximum-likelihood objective:
\begin{equation}
\max_{\theta} \;\; \mathbb{E}_{\mathbf{z} \sim \mathcal{B},\; \mathbf{a}_0 \sim \pi_{\text{lazy}}(\cdot|\mathbf{z})}\!\left[ w(\mathbf{z},\mathbf{a}_0)\, \log p_\theta(\mathbf{a}_0|\mathbf{z}) \right], \quad w(\mathbf{z},\mathbf{a}_0) = \exp\!\left(\frac{A(\mathbf{z},\mathbf{a}_0)}{\eta}\right).
\label{eq:app_weighted_mle}
\end{equation}

When $p_\theta$ is parameterized as a diffusion policy, the marginal log-likelihood $\log p_\theta(\mathbf{a}_0|\mathbf{z})$ becomes intractable. Following the DDPM framework~\citep{ho2020ddpm} (see Section~\ref{subsec:ddpm}), we instead optimize its evidence lower bound (ELBO), whose simplified form reduces to a standard noise-prediction objective. Since the advantage weight $w(\mathbf{z},\mathbf{a}_0) \geq 0$ depends only on the latent state and the clean action, and is independent of both the diffusion timestep $\tau$ and the sampled noise $\boldsymbol{\epsilon}$, it can be taken outside the denoising expectation. Applying this reweighting to Eq.~\ref{eq:app_weighted_mle}, and replacing the theoretical advantage $A$ with its bootstrap estimate $\hat{A}$ from Section~\ref{subsec:dpmpc}, yields the following advantage-weighted score-matching objective:
\begin{equation}
\begin{gathered}
\mathcal{L}_{\mathrm{awr}}(\theta) = \mathbb{E}_{\substack{\tau \sim \mathrm{Unif}\{1,\dots,T\},\; \mathbf{z} \sim \mathcal{B},\; \mathbf{a}_0 \sim \pi_{\text{lazy}}(\cdot|\mathbf{z}),\\ \boldsymbol{\epsilon} \sim \mathcal{N}(\mathbf{0}, \mathbf{I})}} \left[ w(\mathbf{z}, \mathbf{a}_0) \left\| \boldsymbol{\epsilon} - \boldsymbol{\epsilon}_\theta\!\left(\sqrt{\bar{\alpha}_\tau}\, \mathbf{a}_0 + \sqrt{1 - \bar{\alpha}_\tau}\, \boldsymbol{\epsilon},\; \tau,\; \mathbf{z}\right) \right\|^2 \right], \\
w(\mathbf{z}, \mathbf{a}_0) = \exp\!\left(\frac{\hat{A}(\mathbf{z},\mathbf{a}_0)}{\eta}\right).
\end{gathered}
\label{eq:app_awr_loss}
\end{equation}
Applying Eq.~\ref{eq:app_awr_loss} along $H$-step latent rollouts with temporal weighting $\lambda^t$, and normalizing the advantage by the running scale $S$, yields the final DPMPC loss in Eq.~\ref{eq:final_diffusion_loss}.

\section{Experimental Details}
\label{app:exp_details}

\subsection{Benchmarks}

We evaluate on 14 challenging high-dimensional continuous control tasks spanning three benchmark suites: the Dog-Run, Dog-Walk, Dog-Trot, and Dog-Stand environments from DMControl~\citep{tassa2018dmcontrol}; the Key-Turn-Hard, Pen-Twirl-Hard, and Obj-Hold-Hard dexterous manipulation environments from MyoSuite~\citep{MyoSuite2022}; and the H1hand-Run, H1hand-Walk, H1hand-Slide, H1hand-Sit-Hard, H1hand-Stand, H1hand-Pole, and H1hand-Hurdle locomotion environments from HumanoidBench~\citep{sferrazza2024humanoidbench}.
The observation and action space dimensions of the respective environments are shown in Table~\ref{tab:env_dims}.

\begin{table}[htbp]
\centering
\caption{\textbf{Environment specifications.}
Observation and action space dimensions for all evaluation environments.}
\label{tab:env_dims}
\small
\begin{tabular}{lcc}
\toprule
\textbf{Environment} & \textbf{Obs.\ Dim.} & \textbf{Act.\ Dim.} \\
\midrule
Dog-Run          & 223 & 38 \\
Dog-Walk         & 223 & 38 \\
Dog-Trot         & 223 & 38 \\
Dog-Stand        & 223 & 38 \\
\midrule
Key-Turn-Hard    & 93  & 39 \\
Pen-Twirl-Hard   & 83  & 39 \\
Obj-Hold-Hard    & 91  & 39 \\
\midrule
H1hand-Run       & 151 & 61 \\
H1hand-Walk      & 151 & 61 \\
H1hand-Slide     & 151 & 61 \\
H1hand-Sit-Hard  & 151 & 61 \\
H1hand-Stand     & 151 & 61 \\
H1hand-Pole      & 151 & 61 \\
H1hand-Hurdle    & 151 & 61 \\
\bottomrule
\end{tabular}
\end{table}

\subsection{Implementation Details and Hyperparameters}

Our implementation is based on the TD-MPC2~\citep{hansen2023tdmpc2} and BMPC~\citep{wang2025bootstrapped} codebases. For the shared world model components (encoder, latent dynamics, reward predictor, value function, and MPPI planner), we adopt the TD-MPC2 hyperparameter settings without additional tuning and use the same configuration across all tasks. For the value function, we model the state value instead of the state-action value to reduce the number of parameters.
Table~\ref{tab:hyperparams_diffusion} lists the hyperparameters for diffusion policy training introduced in DPMPC.
These values are shared across all tasks.

All DPMPC results are averaged over four random seeds per task. For the BMPC and BOOM baselines, we train and evaluate using their respective official implementations\footnote{\raggedright BMPC codebase: \url{https://github.com/wertyuilife2/bmpc}; BOOM codebase: \url{https://github.com/molumitu/BOOM_MBRL}.}.
For SDAC, we adapt the code from its official repository\footnote{\raggedright SDAC codebase: \url{https://github.com/mahaitongdae/diffusion_policy_online_rl}.} to each task and run it ourselves to obtain the training curves.
For SAC, TD-MPC2, and DreamerV3, we directly adopt the published results from the official TD-MPC2 benchmark\footnote{Baseline results and codebase available at \url{https://github.com/nicklashansen/tdmpc2}.} on the DMControl and MyoSuite tasks, and from the official HumanoidBench benchmark\footnote{Baseline results obtained from \url{https://github.com/carlosferrazza/humanoid-bench}.} on the H1hand tasks.

\begin{table}[htbp]
\centering
\caption{\textbf{Diffusion prior hyperparameters.}
All values are shared across tasks.}
\label{tab:hyperparams_diffusion}
\small
\begin{tabular}{lc}
\toprule
\textbf{Hyperparameter} & \textbf{Value} \\
\midrule
Diffusion steps ($T$) & 20 \\
No. of hidden layers & 2 \\
Hidden dimensions & 256 \\
Activation & Mish \\
Time embedding dimension & 32 \\
Learning rate & $3 \times 10^{-4}$ \\
Advantage-based reweighting clamp range & $[0, 50]$ \\
Reanalyze interval ($K$) & 10 \\
Reanalyze batch size ($B_r$) & 20 \\
Number of MPC realizations per state ($M$) & 2  \\
Advantage temperature ($\eta$) & 0.5 \\
Planning horizon ($H$) & 3 \\
Planner action samples per state & 20 \\
Update-to-data ratio & 1 \\
\bottomrule
\end{tabular}
\end{table}

\subsection{Value Estimation Error Protocol}
\label{app:value_error_protocol}

For each checkpoint, we execute the planner and record 50 states along its trajectories at intervals of 20 environment steps, resetting the environment whenever an episode terminates.
Throughout, the planner runs in exploration mode: each executed action is sampled from the optimized Gaussian $\mathcal{N}(\boldsymbol{\mu}_0^{*}, (\boldsymbol{\sigma}_0^{*})^2)$ rather than set to its mean.
For each recorded state, we reset the simulator to that state and follow the planner until episode termination, computing the discounted return.
We repeat this rollout 50 times per state and average the returns to obtain the reference value $G^{\pi}$.
We report the relative mean absolute error $100\,\sum_i \lvert \hat V(\mathbf{s}_i) - G^{\pi}(\mathbf{s}_i)\rvert / \sum_i G^{\pi}(\mathbf{s}_i)$ over the recorded states.

\section{Additional Ablation Studies}
\label{app:additional_ablations}

\begin{figure}[t]
\centering
\includegraphics[width=0.35\textwidth]{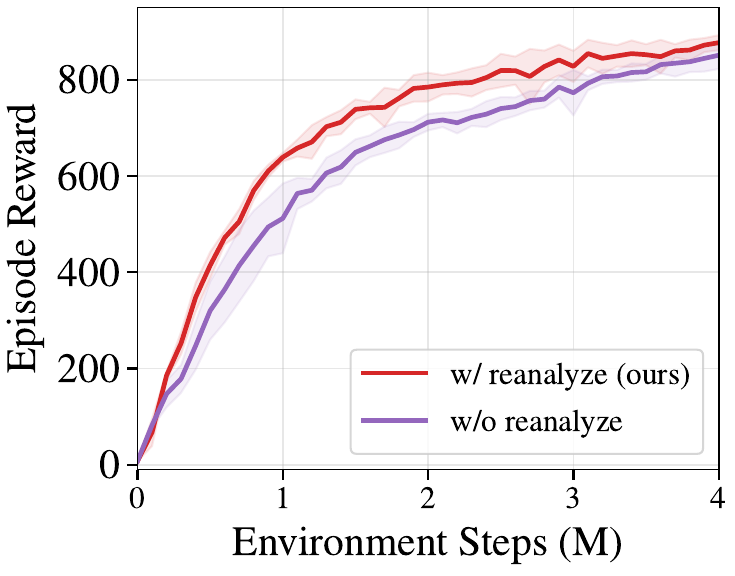}
\caption{\textbf{Ablation of generalized lazy reanalyze on Dog-Run.}
Disabling the periodic refresh of the stored $\pi_{\text{lazy}}$ targets slows learning.}
\label{fig:ablation_reanalyze}
\end{figure}

\paragraph{Effect of generalized lazy reanalyze.}
Figure~\ref{fig:ablation_reanalyze} isolates the contribution of generalized lazy reanalyze by disabling the periodic refresh, so that each stored $\pi_{\text{lazy}}$ target is never recomputed after its first MPC call.
Without the refresh, learning is markedly slower.
This indicates that periodically rerunning MPC is necessary to keep the estimate of the planner distribution up to date.

\begin{figure}[t]
\centering
\begin{subfigure}[b]{0.32\textwidth}
    \centering
    \includegraphics[width=\textwidth]{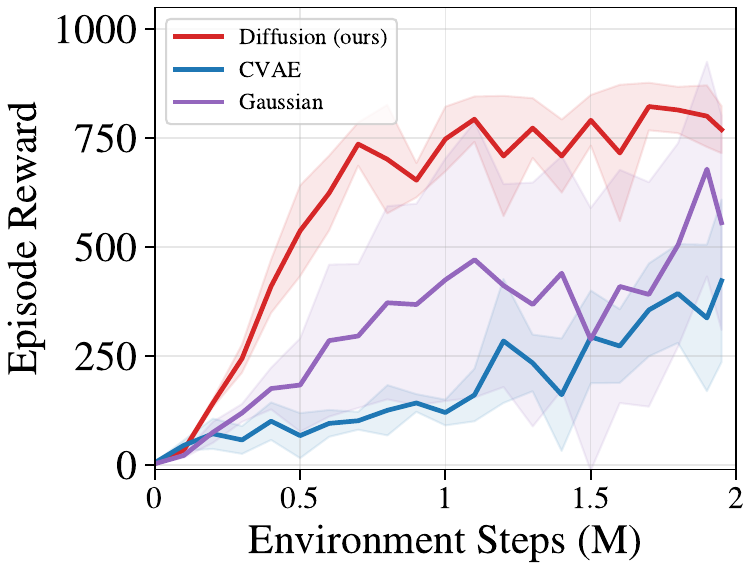}
    \caption{}
    \label{fig:ablation_h1_policy}
\end{subfigure}
\hfill
\begin{subfigure}[b]{0.32\textwidth}
    \centering
    \includegraphics[width=\textwidth]{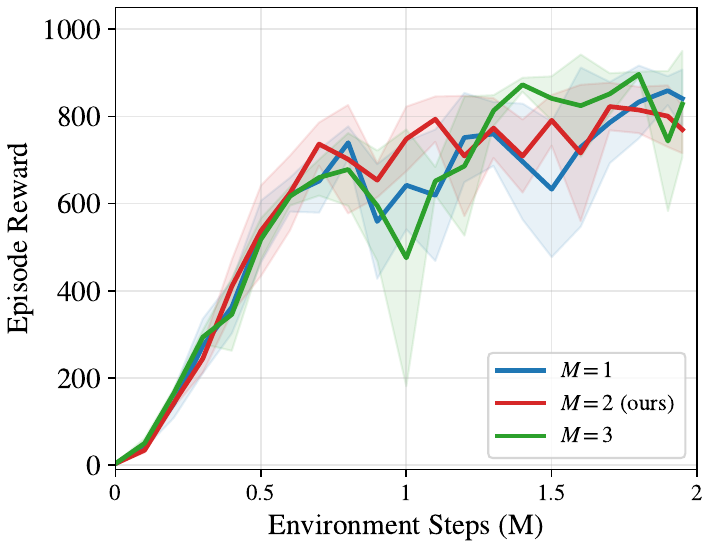}
    \caption{}
    \label{fig:ablation_h1_M}
\end{subfigure}
\hfill
\begin{subfigure}[b]{0.32\textwidth}
    \centering
    \includegraphics[width=\textwidth]{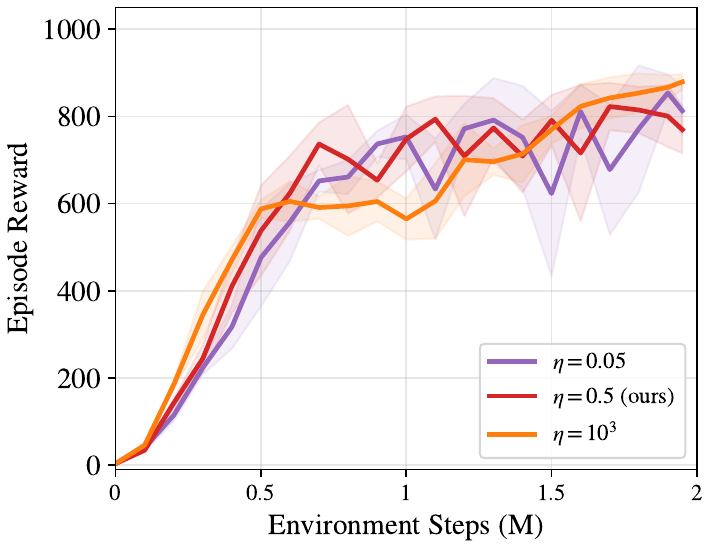}
    \caption{}
    \label{fig:ablation_h1_eta}
\end{subfigure}
\caption{\textbf{Ablation studies on H1hand-Run.}
(\subref{fig:ablation_h1_policy}) Policy prior parameterization.
(\subref{fig:ablation_h1_M}) Number of MPC realizations stored per state $M$.
(\subref{fig:ablation_h1_eta}) Advantage temperature $\eta$.}
\label{fig:ablation_h1}
\end{figure}

\paragraph{Ablations on H1hand-Run.}
Figure~\ref{fig:ablation_h1} repeats the ablations of the policy prior parameterization, the number of MPC realizations $M$, and the advantage temperature $\eta$ on H1hand-Run.
The policy parameterization matters considerably more on H1hand-Run than on Dog-Run (Figure~\ref{fig:ablation_h1_policy}).
The diffusion policy reaches a return of about 750 within 0.7M steps, whereas the Gaussian policy improves slowly with large variance and the CVAE remains below 450 throughout training.
This wider gap on the task with the higher-dimensional action space is consistent with the value estimation analysis in Section~\ref{subsec:value_analysis}, where the advantage of DPMPC is also largest on H1hand-Run.
All three values of $M$ reach a similar final return.
With $M=2$, learning is steadier in the intermediate phase, whereas $M=1$ and $M=3$ show temporary drops around 1M steps.
Setting $\eta=10^3$, which makes the weights nearly uniform, slows learning between 0.7M and 1.2M steps but attains a comparable final return.
Overall, DPMPC is robust to both hyperparameters on H1hand-Run, and the shared defaults ($M=2$, $\eta=0.5$) remain effective on this task.

\section{Additional Visualizations}
\label{app:additional_viz}

\begin{figure}[t]
\centering
\begin{subfigure}[b]{0.49\textwidth}
    \centering
    \includegraphics[width=\textwidth]{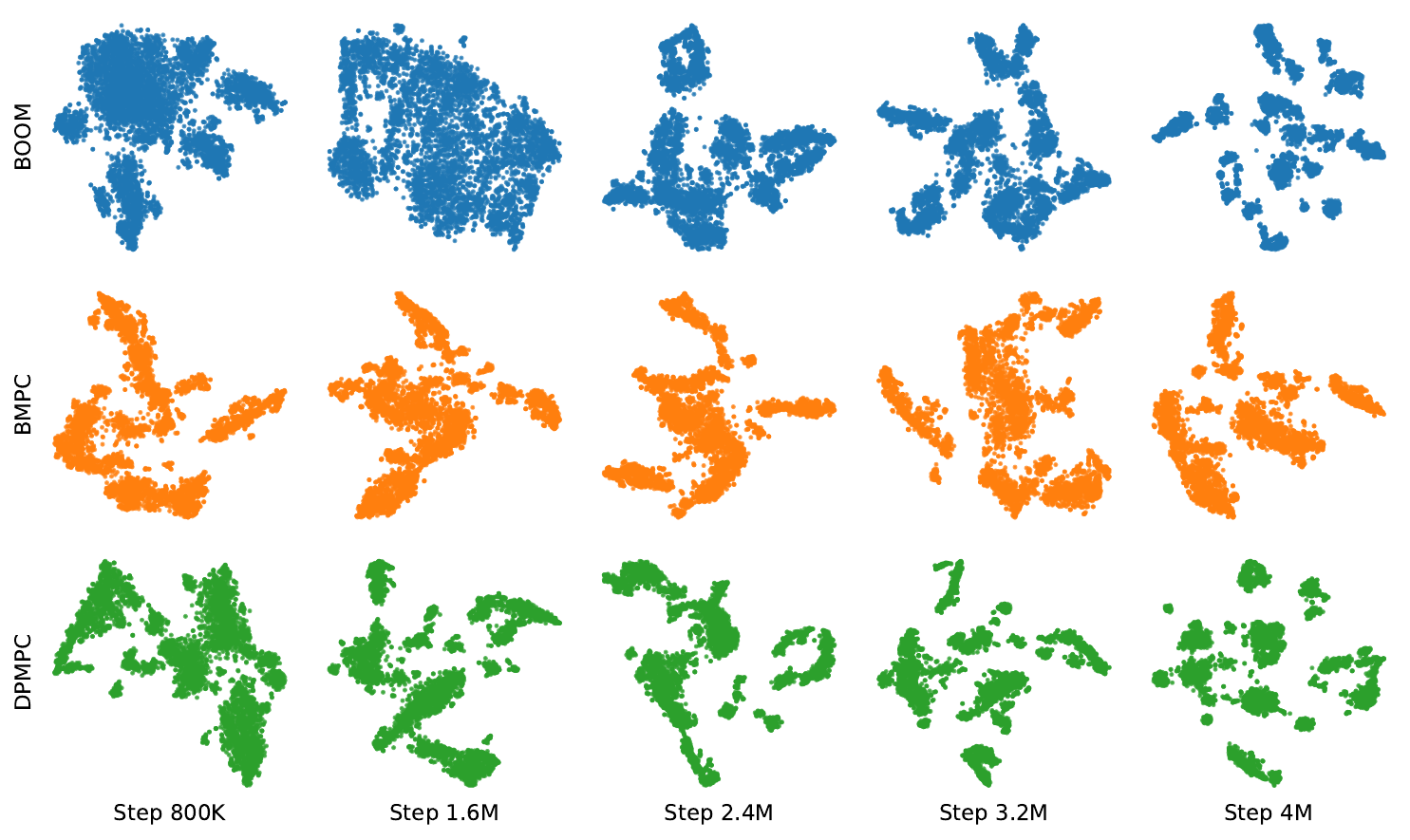}
    \caption{}
    \label{fig:tsne_dog}
\end{subfigure}
\hfill
\begin{subfigure}[b]{0.49\textwidth}
    \centering
    \includegraphics[width=\textwidth]{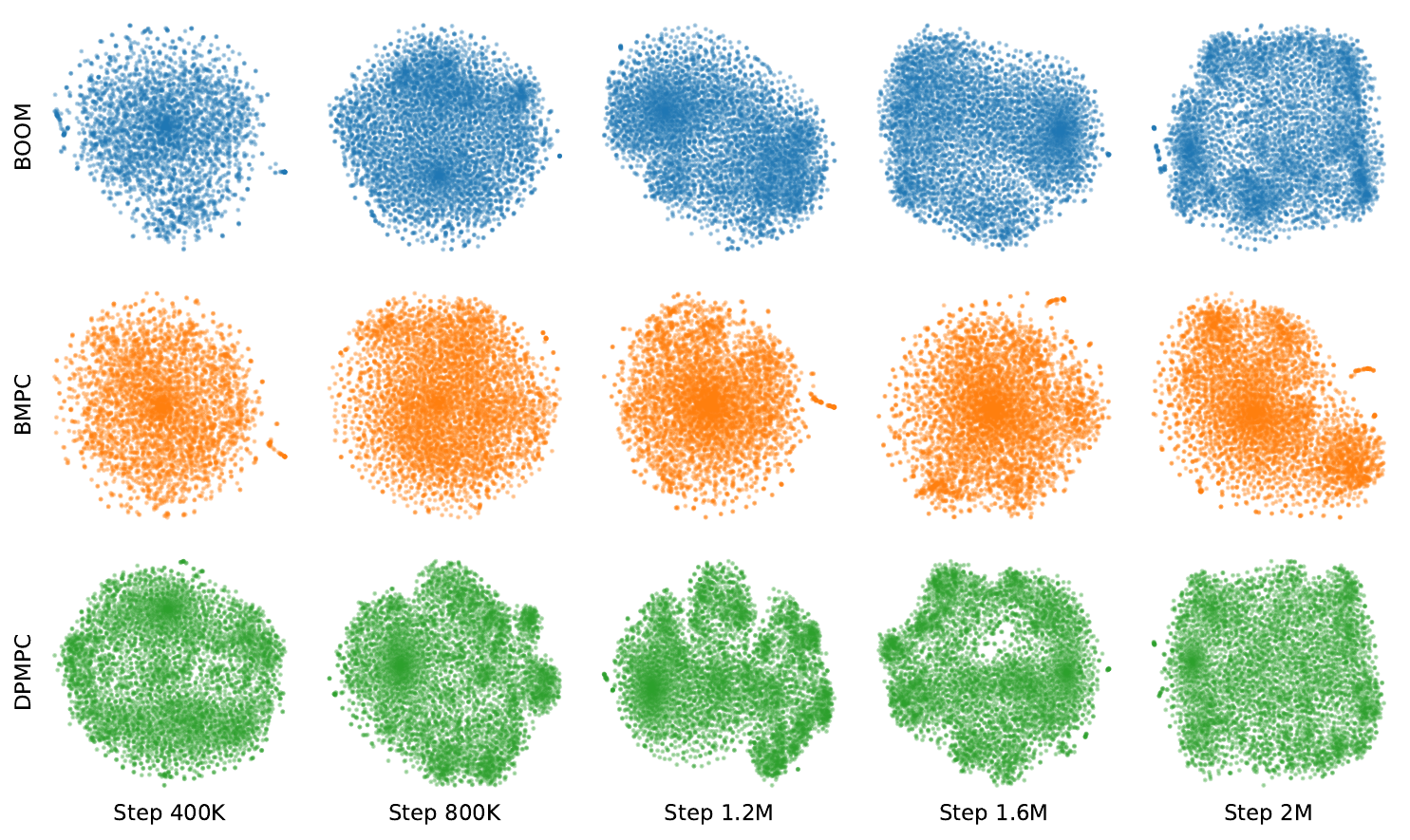}
    \caption{}
    \label{fig:tsne_h1}
\end{subfigure}
\caption{\textbf{State visitation on Dog-Run (\subref{fig:tsne_dog}) and H1hand-Run (\subref{fig:tsne_h1}).}
t-SNE visualization of state visitation distributions across training for DPMPC, BMPC, and BOOM (5 checkpoints, 10 rollout episodes each).
DPMPC visits a substantially more diverse set of states on Dog-Run and H1hand-Run.}
\label{fig:tsne}
\end{figure}

\paragraph{State visitation on Dog-Run and H1hand-Run.}
We examine what a diffusion policy prior changes about the planner's behavior.
We compare the state visitation distributions of DPMPC, BMPC, and BOOM on Dog-Run and H1hand-Run throughout training.
BMPC and BOOM use Gaussian policy priors, whereas DPMPC uses a diffusion policy prior that can represent richer action distributions.
We evaluate five evenly spaced checkpoints and, at each checkpoint, collect 10 rollout episodes before applying t-SNE to project the visited states into two dimensions.
Figure~\ref{fig:tsne_dog} shows that on Dog-Run, DPMPC covers a substantially broader region of the projected state space during the early training phase, even though all three methods achieve comparable episodic returns.
This behavior is consistent with the diffusion prior providing more diverse candidate actions to the planner.
As training progresses, the state visitation distributions of DPMPC and BOOM grow more similar, while BMPC converges to a markedly less diverse region of the state space---consistent with its comparatively weaker asymptotic performance.
Figure~\ref{fig:tsne_h1} provides the corresponding t-SNE visualization on H1hand-Run.
Consistent with our Dog-Run findings, the low-dimensional state distribution exhibits multiple modes throughout training.
The number of modes is noticeably larger than for the two baselines, suggesting more diverse behavior under DPMPC.

\begin{figure}[htbp]
\centering
\includegraphics[width=\textwidth]{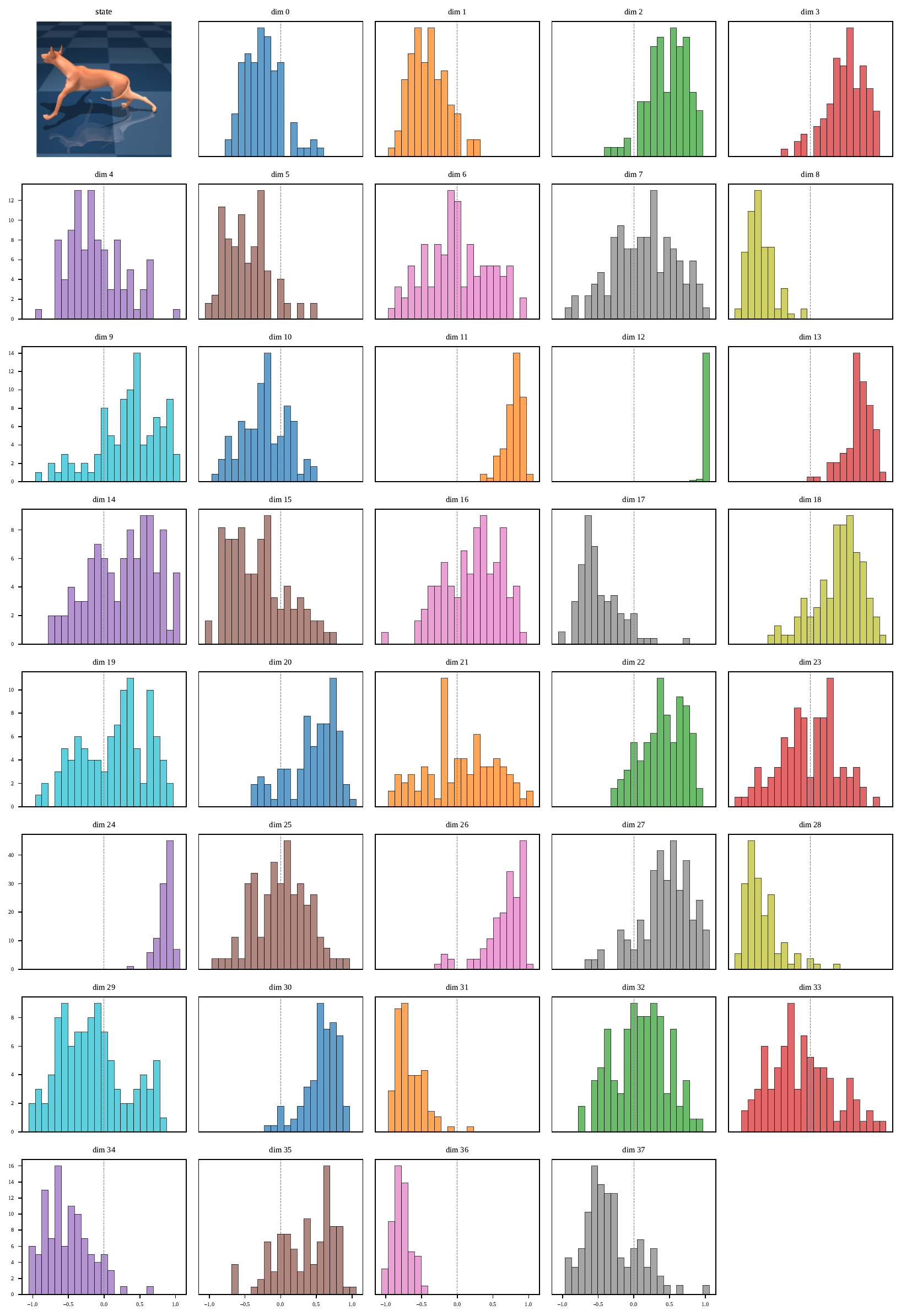}
\caption{\textbf{Action distributions on Dog-Run.}
Per-dimension action histograms sampled from the learned diffusion policy (2M steps).
Multiple dimensions display multimodal or non-Gaussian distributions, demonstrating the expressiveness of the diffusion policy prior.}
\label{fig:diffusion_histograms_dog}
\end{figure}

\begin{figure}[htbp]
\centering
\includegraphics[width=\textwidth]{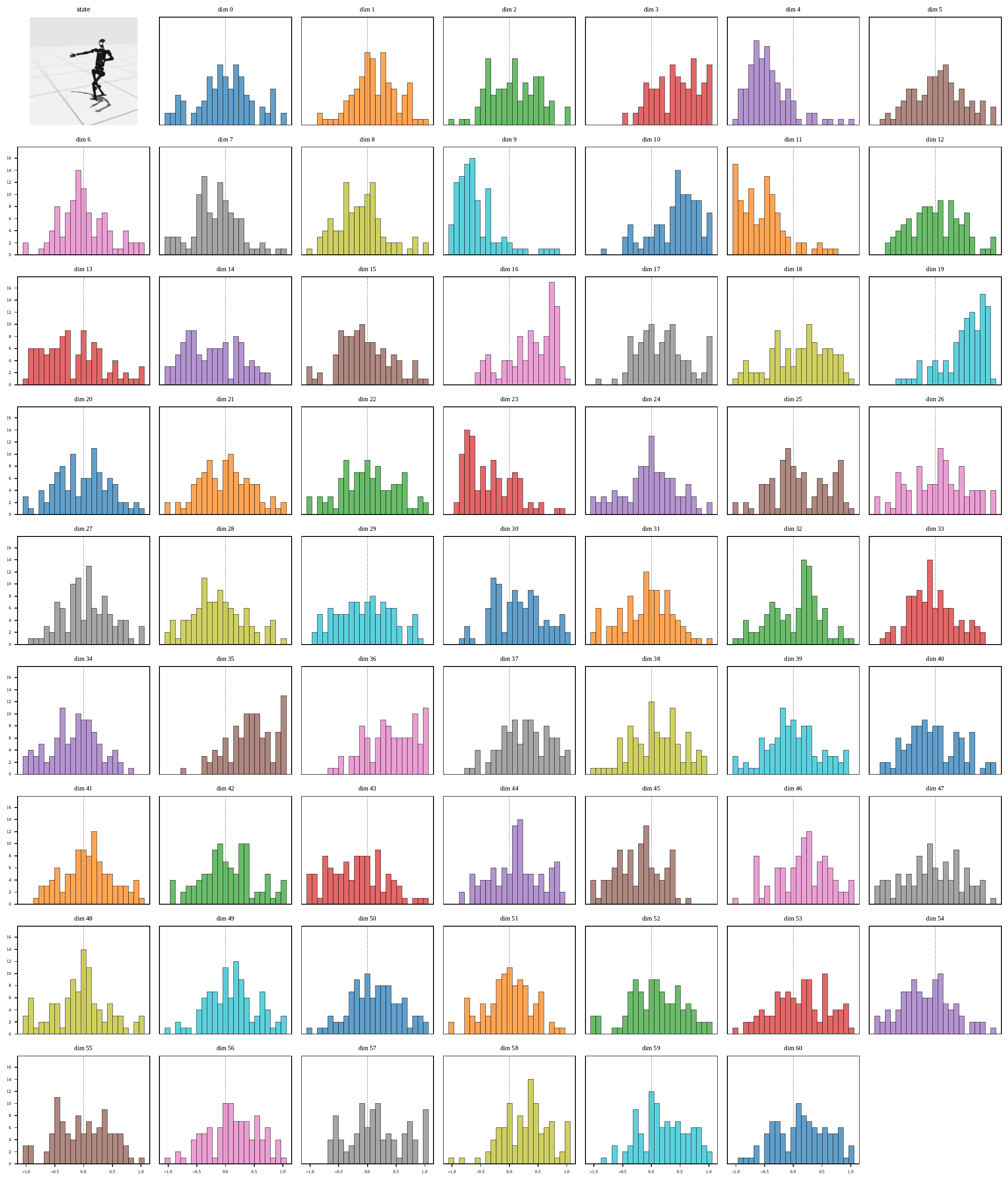}
\caption{\textbf{Action distributions on H1hand-Run.}
Per-dimension action histograms sampled from the learned diffusion policy (1M steps).}
\label{fig:diffusion_histograms_h1hand}
\end{figure}

\paragraph{Diffusion policy action distribution.}
Figures~\ref{fig:diffusion_histograms_dog} and~\ref{fig:diffusion_histograms_h1hand} visualize the per-dimension action distributions sampled from the learned diffusion policy on Dog-Run (checkpoint at 2M steps) and H1hand-Run (checkpoint at 1M steps), respectively.
Unlike a Gaussian policy, which constrains each action dimension to a unimodal distribution, the diffusion policy captures multimodal and non-Gaussian structure.
This expressiveness is evident across both tasks: many dimensions exhibit clearly separated modes or heavy-tailed shapes that a Gaussian prior could not represent, providing a richer set of candidate actions for the model-based planning stage.

\section{Training Time Comparison}
\label{app:training_time}

Table~\ref{tab:training_time} reports the wall-clock training time on the Dog-Run and H1hand-Run tasks over 2M environment steps.
Dog-Run uses an action repeat of 2, and H1hand-Run uses an action repeat of 1.
All methods are benchmarked on the same hardware: a single NVIDIA RTX 3090 GPU with an AMD Ryzen Threadripper 3970X CPU, and each method uses a single GPU for training.
The results show a moderate increase in training time compared with TD-MPC2 and BMPC, while DPMPC remains faster than BOOM.
This overhead could be reduced by faster samplers~\citep{song2020denoising} and alternative generative-policy parameterizations, such as consistency models~\citep{song2023consistency} or flow-matching models~\citep{lipman2022flow}.

\section{Action Selection of MPC}
\label{app:mppi_bimodal}

\begin{figure}[htbp]
\centering
\includegraphics[width=\textwidth]{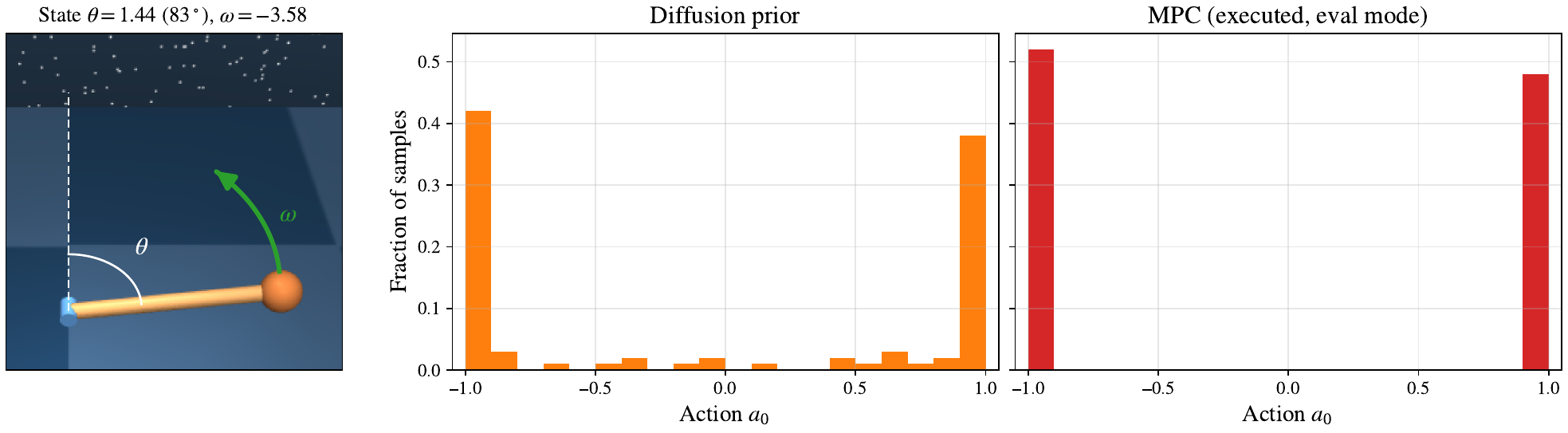}
\caption{\textbf{Prior and executed MPC actions at a bimodal pendulum state.}
Left: the selected state.
Middle: 100 actions $a_0$ sampled from the diffusion prior.
Right: 100 actions executed by MPC in evaluation mode.}
\label{fig:mppi_bimodal_dist}
\end{figure}

\begin{figure}[htbp]
\centering
\includegraphics[width=\textwidth]{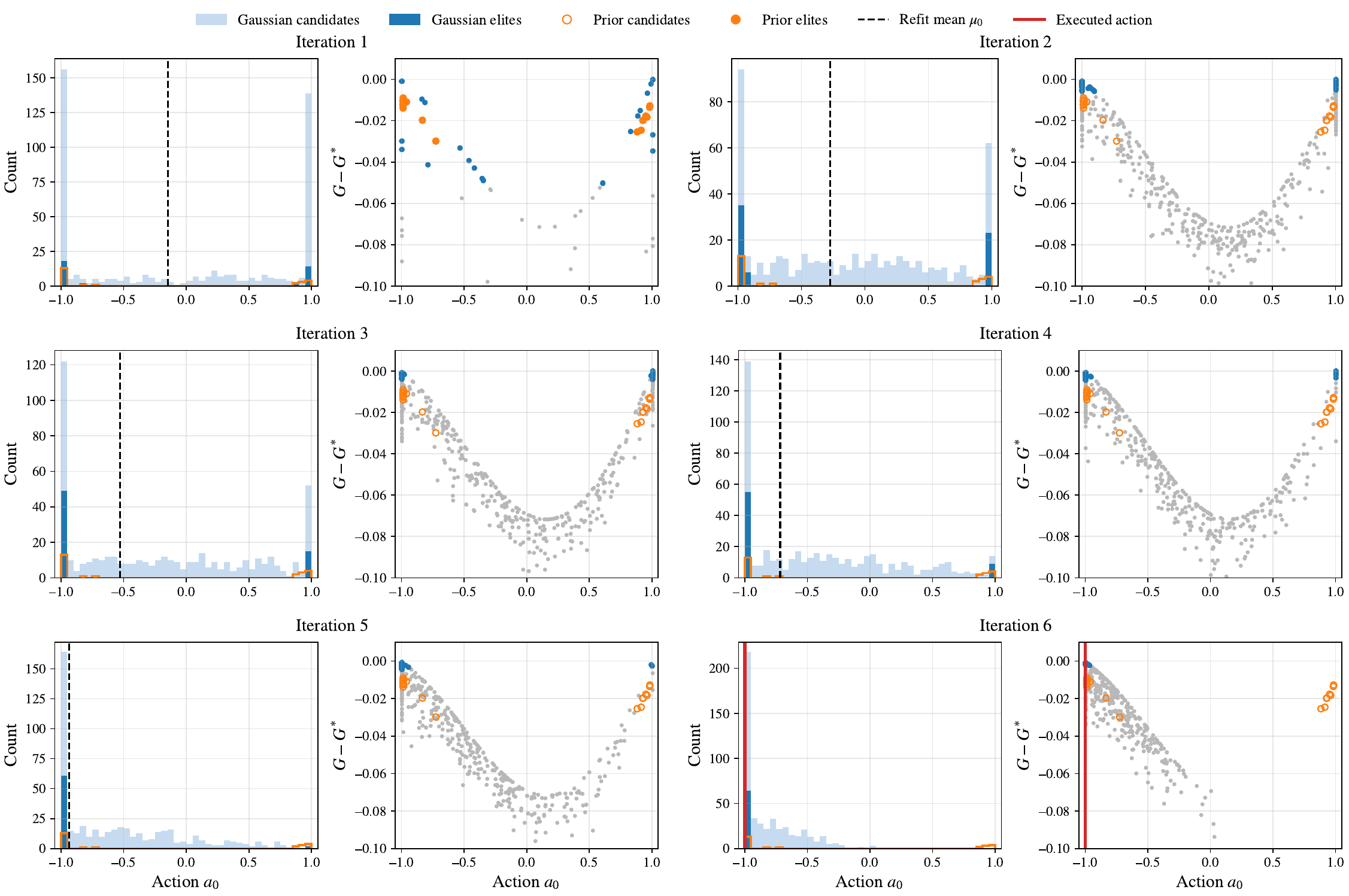}
\caption{\textbf{Evolution of a single MPPI planning call at the same state.}
For each iteration, the left panel shows the action $a_0$ of the Gaussian and prior candidates and elites, with the refitted mean $\mu_0$ (dashed).
The right panel plots the return gap $G-G^*$ of each candidate against $a_0$, where $G^*$ is the maximum return among all candidates (grey: non-elite candidates).
The red line in iteration 6 marks the executed action.}
\label{fig:mppi_bimodal_trace}
\end{figure}

We illustrate the action selection described in Section~\ref{subsec:planning_based_mbrl} on the DMControl Pendulum-Swingup task with a trained DPMPC agent.
We select a state (Figure~\ref{fig:mppi_bimodal_dist}, left) at which the learned diffusion prior places probability mass on both action bounds, and both bounds attain nearly the same return.
Figure~\ref{fig:mppi_bimodal_dist} compares 100 actions $a_0$ sampled from the prior with 100 actions executed by MPC in evaluation mode at this state.
The executed actions split almost evenly between the two modes (52\% vs.\ 48\%), with no compromise action between them.
Figure~\ref{fig:mppi_bimodal_trace} traces a single planning call at the same state.
In early iterations, the refitted mean $\mu_0$ lies between the two modes, where candidates attain the lowest return.

\section{Action Distributions of CVAE and Diffusion Priors}
\label{app:prior_hist}
\begin{figure}[htbp]
\centering
\begin{subfigure}[b]{\textwidth}
    \centering
    \includegraphics[width=\textwidth]{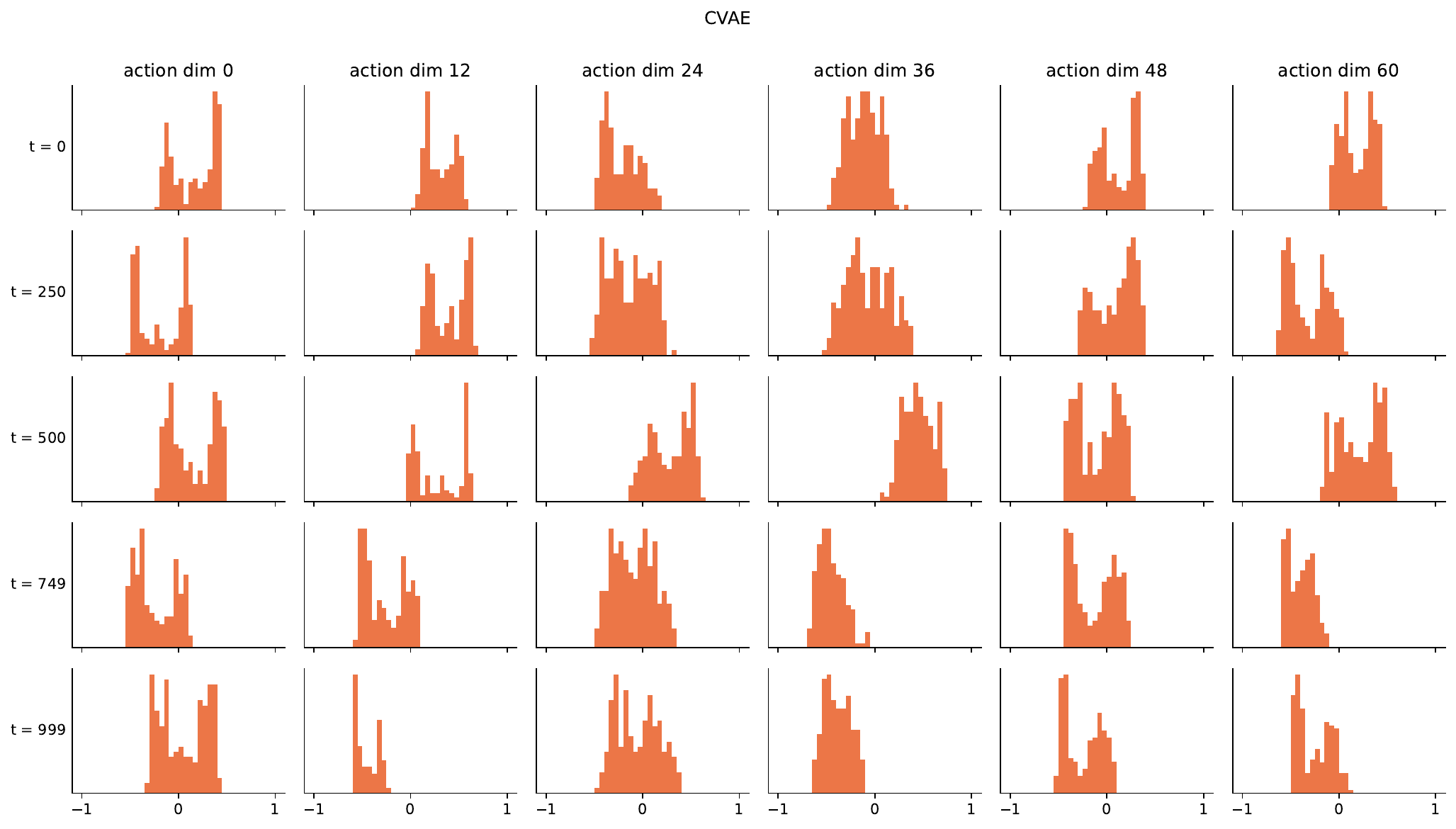}
    \caption{}
    \label{fig:prior_hist_cvae}
\end{subfigure}
\vspace{0.5em}
\begin{subfigure}[b]{\textwidth}
    \centering
    \includegraphics[width=\textwidth]{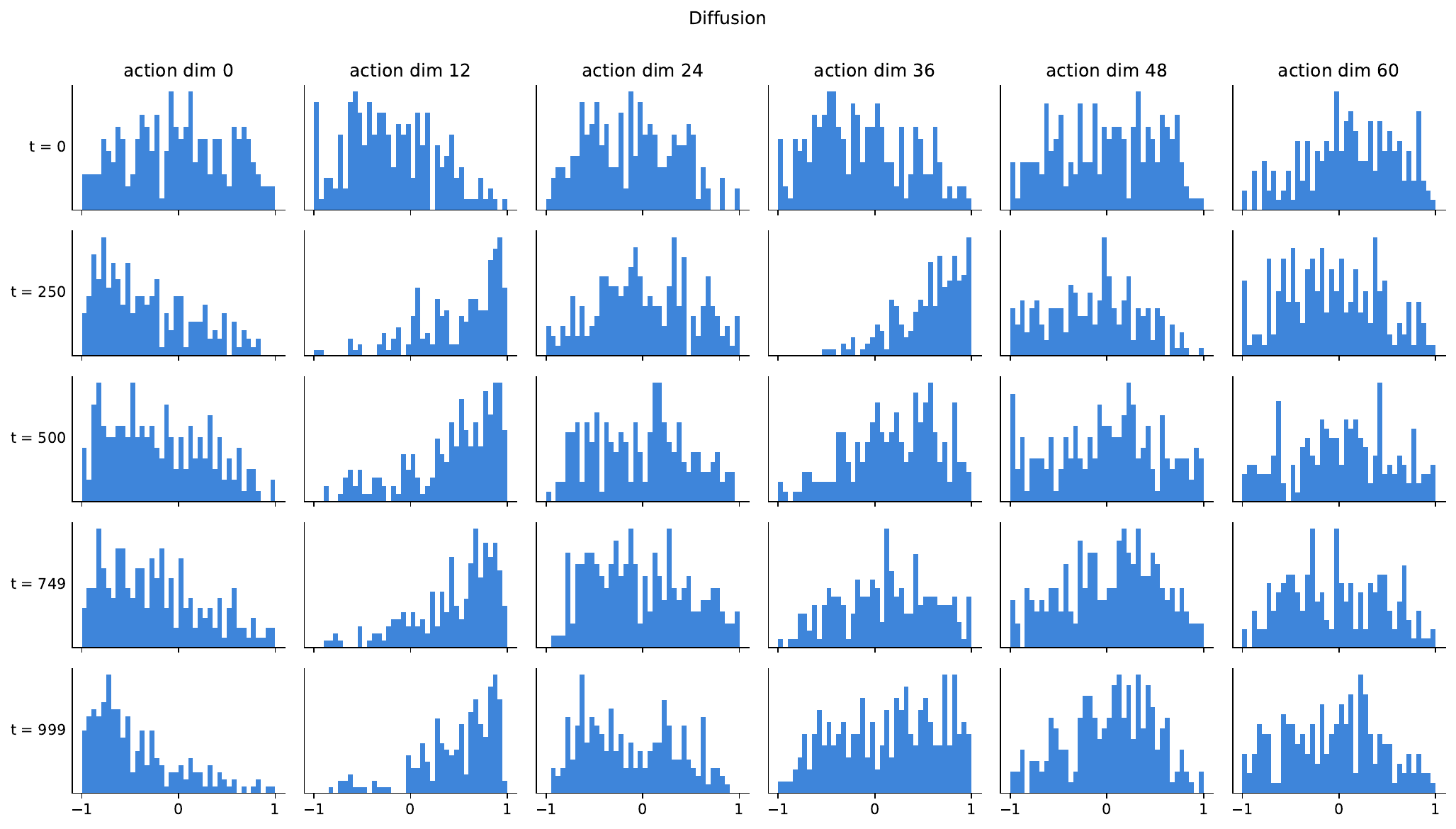}
    \caption{}
    \label{fig:prior_hist_diffusion}
\end{subfigure}
\caption{\textbf{Action distributions of the CVAE prior (\subref{fig:prior_hist_cvae}) and the diffusion prior (\subref{fig:prior_hist_diffusion}) on H1hand-Run.}
Each panel shows the histogram of 200 actions sampled from the prior at the state reached at step $t$ of one episode.}
\label{fig:prior_hist}
\end{figure}

To analyze the effect of expressive policy class, we simply replace the diffusion prior as CVAE model.  
We compare the action distributions learned by the two prior on H1hand-Run.
For each prior, we load the final checkpoint and roll out one episode of 1000 steps in the simulator.
We select five evenly spaced states from this episode ($t \in \{0, 250, 500, 749, 999\}$) and draw 200 action samples from the prior at each state.
Figure~\ref{fig:prior_hist} shows the resulting histograms for six action dimensions.

The two priors learn markedly different action distributions.
The CVAE samples concentrate in a narrow interval and mostly form two clusters.
The diffusion samples cover nearly the full action range $[-1, 1]$, and their shape varies with both the state and the action dimension.
This indicates that the CVAE prior learns a simpler action distribution, whereas the diffusion prior captures richer structure in the action space.

\end{document}

%% file: math_commands.tex
\usepackage{amsmath,amsfonts,bm}

\def\eqref#1{equation~\ref{#1}}

\def\1{\bm{1}}

\DeclareMathAlphabet{\mathsfit}{\encodingdefault}{\sfdefault}{m}{sl}
\SetMathAlphabet{\mathsfit}{bold}{\encodingdefault}{\sfdefault}{bx}{n}

